\pdfoutput=1

\documentclass[11pt]{article}

\usepackage{EMNLP2023}
\usepackage{tabularx}
\usepackage{booktabs, multirow} 
\usepackage{soul}
\usepackage{float}
\restylefloat{table}
\usepackage{subfigure}
\usepackage{times}
\usepackage{latexsym}
\usepackage{tikz}
\usepackage{amsfonts}
\usepackage{amsmath}
\usepackage{xspace}
\usepackage{enumitem}
\usetikzlibrary{topaths}
\usepackage{comment}
\tikzstyle{every picture}+=[remember picture,inner xsep=0,inner ysep=0.05ex]

\usepackage[utf8]{inputenc}

\usepackage{microtype}

\usepackage{inconsolata}
\usepackage{fontawesome}

\usepackage{eccvabbrv}

\usepackage{graphicx}
\usepackage{booktabs}

\usepackage[accsupp]{axessibility}  
\usepackage{xcolor,colortbl}
\definecolor{myblue}{HTML}{d8effd}
\definecolor{mypurple}{HTML}{e2d7ed}

\usepackage{hyperref}

\usepackage{orcidlink}

\newcommand{\replace}{\texttt{REPLACE}\xspace}
\newcommand{\swap}{\texttt{SWAP}\xspace}

\begin{document}

\title{The Hard Positive Truth about Vision-Language Compositionality} 


\author{
Amita Kamath$^{1,2}$ \quad
Cheng-Yu Hsieh$^1$ \quad
Kai-Wei Chang$^2$ \quad
Ranjay Krishna$^{1,3}$\\
$^1$ University of Washington \\
$^2$ University of California, Los Angeles \\
$^3$ Allen Institute for AI \\
{  \normalsize \url{https://github.com/amitakamath/hard_positives}} 
}

\twocolumn[{%
\renewcommand\twocolumn[1][]{#1}%
\maketitle
\begin{center}
    \centering
    \includegraphics[width=\textwidth]{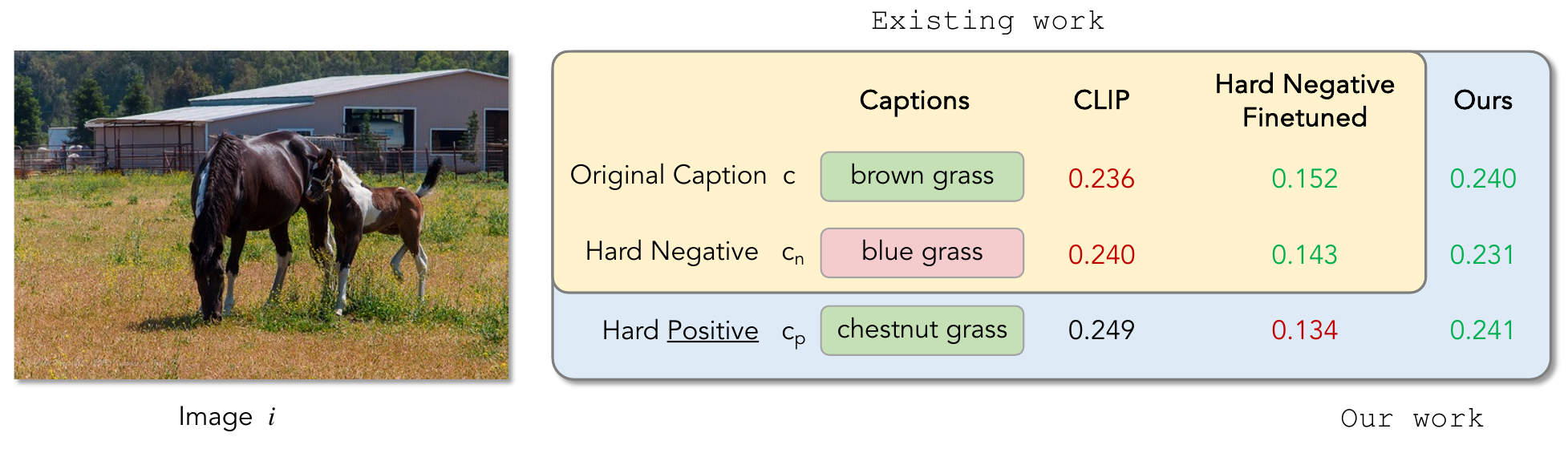}
    \captionof{figure}{
  Prior work shows that CLIP is insensitive to minor changes to the input caption, incorrectly assigning a higher score to a hard negative caption $c_n$ than to the original caption $c$. 
  While hard negative finetuning (here, \citet{doveh2023dense}) fixes the ordering between the original caption and the hard negative, we reveal that the resulting model becomes oversensitive and incorrectly assigns a lower score to a hard \textit{positive} caption $c_p$. We mitigate this by finetuning with both hard negatives and hard positives, leading to an overall correct understanding of the different captions, and achieving a more well-rounded sense of compositionality (real example shown). \\}
  \label{fig:teaser}
  \end{center}%
}]

\begin{abstract}
    Several benchmarks have concluded that our best vision-language models (\eg, CLIP) are lacking in compositionality.
Given an image, these benchmarks probe a model's ability to identify its associated caption amongst a set of compositional distractors.
In response, a surge of recent proposals show improvements by finetuning CLIP with distractors as \textbf{hard negatives}.
Our investigations reveal that these improvements have, in fact, been significantly overstated --- because existing benchmarks do not probe whether finetuned vision-language models remain invariant to \textbf{hard positives}\footnote{Correspondence to: \texttt{kamatha@cs.washington.edu} }.
By curating an evaluation dataset with $112,382$ hard negatives and hard positives, we uncover that including hard positives decreases CLIP's performance by $12.9\%$, while humans perform effortlessly at $99\%$. CLIP finetuned with hard negatives results in an even larger decrease, up to $38.7\%$. 
With this finding, we then produce a 1,775,259 image-text training set with both hard negative and hard positive captions. 
By training with both, we see improvements on existing benchmarks while simultaneously improving performance on hard positives, indicating a more robust improvement in compositionality.
Our work suggests the need for future research to rigorously test and improve CLIP's understanding of semantic relationships between related ``positive'' concepts. 
\end{abstract}



\section{Introduction}

Compositionality is a fundamental characteristic of both human vision as well as natural language. It suggests that ``the meaning of the whole is a function of the meaning of its parts''\cite{cresswell1973logics}. 
For instance, compositionality allows people to differentiate between a photo of ``a brown dog holding a white frisbee'' and ``a white dog running after a brown frisbee''.
For a while now, research on vision-language models has sought to inject such compositional structure as inductive priors so that models can comprehend scenes and express them using compositional language~\cite{krishna2017visual,ji2020action,lu2016visual,GrundeMcLaughlin2021AGQA}.
However, with the rise of large-scale pretraining, vision-language models today are trained from image-text pairs scraped from the internet~\cite{thomee2016yfcc100m,schuhmann2022laion,sharma2018conceptual}, and thus, are not explicitly given structural priors.

To probe whether large-scale pretrained vision-language models, such as CLIP~\cite{RadfordKHRGASAM21}, are capable of compositional reasoning, a number of contemporary benchmarks have been released~\cite{thrush2022winoground,zhao2022vl,yuksekgonul2023when,ma2022crepe,ray2023cola,hsieh2023sugarcrepe, kamath2023text}.
Evaluation is primarily conducted through an image-to-text retrieval task formulation~\cite{zhao2022vl,yuksekgonul2023when,ma2022crepe}: by measuring how often models pick the description, ``a brown dog holding a white frisbee'' when presented with an image of it, and avoid choosing the incorrect \textbf{hard negative} description, ``a {\color{orange} white} dog {\color{purple}running after} a {\color{orange}brown} frisbee''. This second sentence is considered a hard negative because the colors are {\color{orange}swapped} and the verb is {\color{purple}replaced}.
Surprisingly, these benchmarks unanimously find that state-of-the-art models demonstrate little to no compositionality~\cite{hsieh2023sugarcrepe}.

As a natural follow up, many approaches have been proposed to remedy this lack of compositionality~\cite{zheng2024iterated}. The most common method finetunes the CLIP model with similar hard negatives. Intuition suggests that by exposing CLIP to hard negatives, it will learn when such perturbations change the semantic meaning of the caption, and therefore should be sensitive to them~\cite{yuksekgonul2023when, doveh2023teaching}.
With hard negative finetuning, results on benchmarks appear to suggest that CLIP models become more compositional~\cite{hsieh2023sugarcrepe}. However, our results indicate otherwise.

We create a new evaluation dataset of $56,191$ images with $28,748$ {\color{orange}swap} and $27,443$ {\color{purple}replace} \textbf{hard positives}.
Hard positives, in contrast to their negative counterparts, make semantic-\textit{preserving} changes to concepts in an original caption. For example, ``a brown dog {\color{purple}holding} $\ldots$'' and ``a brown dog {\color{purple}grasping} $\ldots$'' are {\color{purple}replaced} hard positives.
Ideally, models should be invariant to semantics-preserving perturbations.
We validate this evaluation set with a human evaluation, where our participants effortlessly achieved $99\%$.

Our experiments reveal that the default CLIP model~\cite{RadfordKHRGASAM21} performs $14.9\%$ worse on our data versus on existing benchmarks.
Worse, we test $7$ CLIP finetuning approaches~\cite{yuksekgonul2023when,ma2022crepe,hsieh2023sugarcrepe,doveh2023teaching,doveh2023dense} to find even sharper decreases in performance, up to $38.7\%$.
We find that hard negative-finetuned models are ``oversensitive'', \ie, they more often rank hard negatives higher than one but not both the original caption and the hard positive. We summarize these ideas in Figure~\ref{fig:teaser}.

To mitigate oversentitivity and this general degradation of performance, we curate a larger training set of $591,753$ hard positives and explore a simple data-augmentation training technique wherein CLIP models are finetuned simultaneously with both hard negatives and positives, in addition to the original caption. 
Compared to the original CLIP model, exposure to both improves performance in existing benchmarks and our evaluation data.
When compared to models finetuned only on hard negatives, our model retains most of the performance improvements on existing benchmarks while improving on our evaluation set.
We also find that exposure to only {\color{orange}swap} positives mitigates oversensitivity on the {\color{orange}swap} evaluation set and not on {\color{purple}replace} evaluation set, and vice versa.

Taken together, our investigations expose another dimension of compositionality which was previously unexplored by existing benchmarks. We lay out a number of implications of our findings in our discussion. We release our code, datasets and models at \url{https://github.com/amitakamath/hard_positives}.


\section{Related work}
\label{sec:related_work}
We contextualize our study within research aiming to evaluate and improve the compositionality of vision-language models.

\noindent\textbf{Benchmarks for vision-language compositionality.}
There has been a surge of benchmarks to assess how well vision-language models represent compositional concepts~\cite{yuksekgonul2023when,thrush2022winoground,zhao2022vl, ma2022crepe, ray2023cola, hsieh2023sugarcrepe,kamath2023text}. These tools often reveal that, despite achieving impressive results in various applications~\cite{RadfordKHRGASAM21, 0001LXH22, singh2022flava, alayrac2022flamingo, wang2022omnivl,wang2022image, zhai2022lit}, these models struggle with basic compositional tasks. Issues include difficulty in processing sentences with the same words in a different order~\cite{thrush2022winoground}, and in recognizing relationships between objects or associating objects with their attributes~\cite{zhao2022vl, yuksekgonul2023when, ray2023cola,hsieh2023sugarcrepe, bugliarello2023measuring}. 
Benchmarks also reveal that many models struggle with spatial reasoning~\cite{zellers2018swag,parcalabescu-etal-2022-valse,hendricks-nematzadeh-2021-probing,kamath-etal-2023-whats}.
Our evaluation dataset complements these benchmarks by introducing the notion of hard positives which allows us to uncover that hard negative finetuning induces behaviors that bring into question their semantic understanding of concepts.

\noindent\textbf{Hard negative finetuning for compositionality.}
Efforts to bolster the compositional capabilities of vision-language models have introduced strategies that incorporate new data, methodologies, and loss functions~\cite{yuksekgonul2023when, cascantebonilla2023going, ray2023cola, doveh2023teaching, Singh2023CoarsetoFineCL}. A key strategy involves training models to differentiate between correct captions and procedurally-generated hard negatives~\cite{yuksekgonul2023when, doveh2023teaching, doveh2023dense}. 
However, it remains uncertain whether these approaches genuinely foster a deeper understanding of compositionality or merely enable models to perform well on dataset biases~\cite{hsieh2023sugarcrepe}. Our study explores this question to provide evidence that models do in fact \textit{appear} to perform better on existing benchmarks, but produce the undesirable side effect of being overly sensitive even to semantic-preserving perturbations. 

\noindent\textbf{Mitigating biases in datasets.}
The challenge of biased datasets, which can artificially inflate the perceived effectiveness of models, has been well-documented~\cite{gururangan-etal-2018-annotation}. Several studies propose methods for de-biasing these datasets to ensure evaluations more accurately reflect model capabilities~\cite{reif2023fighting,zellers2018swag,sakaguchi2021winogrande,le2020adversarial}. Techniques like adversarial filtering~\cite{zellers2018swag} use a set of classifiers to eliminate easily guessable instances, creating a tougher benchmark. AFLite builds on this by offering a simplified approach to filtering without needing iterative model retraining, leading to benchmarks that more closely align with the intended tasks~\cite{sakaguchi2021winogrande,le2020adversarial}. In the context of vision-language compositionality evaluation, SugarCrepe identifies and fixes several textual biases exhibiting in procedurally-generated hard negatives in prior benchmarks, yet it only uses hard negatives as in prior benchmarks~\cite{hsieh2023sugarcrepe}. We complement these benchmarks by introducing hard positives to allow a comprehensive evaluation of vision-language compositionality.

\noindent\textbf{Augmenting model training with rewritten captions.}
In addition to hard negative mining, several recent works have explored augmenting data with caption-rewriting methods to improve vision-language models' performance~\cite{doveh2023dense, doveh2023teaching, fan2023improving}. These works typically utilize large language models~\cite{chatgpt, workshop2022bloom} to rewrite a given caption into a very different, new caption describing the same scene, in the hope that the generated captions enrich language supervision for model learning. In this work, we show that even by augmenting model training with the rewritten \textit{positive} captions, the oversensitivity introduced by hard negative finetuning~\cite{doveh2023dense, doveh2023teaching} is so dire that models still fail to correctly identify hard positives from negatives. However, we show that by training with \textit{hard} positives, we are able to better mitigate models' oversensitivity issue.


\section{Evaluating for compositionality}

This section formalizes the principle of compositionality to a well-defined evaluation scheme~\cite{hupkes2020compositionality}. First, we establish how vision-language compositionality is defined (Section \ref{sec:definition}). Then, we explain how existing benchmarks evaluate compositionality (Section \ref{sec:protocol}) and their limitations under this definition (Section \ref{sec:limited_eval}). Finally, we explain how we overcome this limitation by developing a new evaluation dataset (Section \ref{sec:benchmarks}).

\subsection{Definition of compositionality}
\label{sec:definition}
To evaluate the compositionality of vision-language models, most existing benchmarks define a compositional language consisting of \textit{scene graph} visual concepts~\cite{ma2022crepe} or a subset of scene graphs (\eg some focus only on spatial relationships~\cite{parcalabescu-etal-2022-valse, kamath-etal-2023-whats}). 
Within this language, an \textit{atom} $a$ is defined as a singular visual concept, corresponding to a single scene graph node. A \textit{compound} $c$ is defined as a primitive composition of multiple atoms, which corresponds to connections between scene graph nodes.
Scene graphs admit two compound types: the attachment of attribute to objects (``brown dog”), and the attachment of two objects via a relationship (``dog runs after frisbee”).

In most cases, we use entire captions to represent compounds $c$ found in existing vision-language datasets. Conversely, captions can be parsed to become scene graphs. 
It has been shown that scene graphs, through this compositional language, are capable of capturing a number of linguistic phenomena~\cite{suhr-etal-2019-corpus,parcalabescu-etal-2022-valse}, including the existence of concepts (``a photo with \textit{dog}''), spatial relationships (``a grill \textit{on the left of} a staircase''), action relationships (``a dog \textit{holding} a frisbee''),  
prepositional attachment (``A \textit{brown} dog''), and negation (``There are \textit{no} cats'').

\subsection{Evaluation protocol}
\label{sec:protocol}
A majority of existing compositionality benchmarks for vision-language models formulate the evaluation task as image-to-text retrieval~\cite{zhao2022vl,yuksekgonul2023when,ma2022crepe}.
Given an image, the model is probed to select text that correctly describes the image from a pool of candidates.
Unlike standard retrieval tasks where the negative (incorrect) candidates differ significantly from the positive (correct) text, compositionality benchmarks intentionally design \textbf{hard negative} texts that differ minimally from the positive text, in order to test whether the model understands the fine-grained atomic concepts that compose the scene. Under the definition above, hard negatives are defined as compounds with an atom either {\color{orange}swapped} or {\color{purple}replaced}. Both operations modify the compound such that their semantic interpretation violates the visual concepts in their corresponding image.

Re-using the example from the introduction, we have an image of ``a brown dog holding a white frisbee''. In comparison, ``a {\color{orange} white} dog {\color{purple}running after} a {\color{orange}brown} frisbee'' is a compound with multiple negative operations. The attributes  {\color{orange} white} and {\color{orange} brown} are {\color{orange}swapped} and the relationship {\color{purple}holding} is {\color{purple}replaced} by {\color{purple}running after}.
Most benchmarks curate evaluation sets with multiple hard negatives per image-text pair.

Using such a benchmark, they define the compositionality evaluation protocol as follows:
Given a query image $i$, the model is tasked with retrieving its corresponding compound caption $c$ amongst a set of distractors. Without loss of generality, assume there is one distractor $c_n$ per image.
The protocol first estimates a matching score between the image and each of the captions (image-text matching score): $s(c,i)$, $s(c_n,i)$. If a model is compositional, $s(c,i) > s(c_n,i)$, resulting in retrieving the correct caption over the hard negative.

\subsection{Limitations with existing evaluations}
\label{sec:limited_eval}
The assumption made by existing benchmarks is that all atomic swaps or replacements necessarily cause a change in semantics. However, this is not the case with language. For example, ``a brown dog {\color{purple}holding} $\ldots$'' and ``a brown dog {\color{purple}grasping} $\ldots$'' are {\color{purple}replaced} hard positives since the {\color{purple}replacement} of {\color{purple}holding} to {\color{purple}grasping} does not alter the caption's grounding with respect to the image.

As such, we posit that existing benchmarks are incomplete. They have left out a vital component of compositionality: \textbf{hard positives}.
Compositional models should be able to reason about two kinds of operations: (1) when a modification to $c$ produces a hard negative $c_n$, the $s(c_n,i)$ should reduce when compared to $s(c,i)$; and (2) when a modification to $c$ produces a hard positive $c_p$, then $s(c_p,i)$ should remain relatively similar to $s(c,i)$. In summary, hard positives should not alter the score $s(c,i) \approx s(c_p,i)$.

\subsection{Curating a hard positive evaluation dataset}
\label{sec:benchmarks}
We respond to this incomplete evaluation by curating  an evaluation dataset with hard positives. We focus on the two main types of perturbations in existing work: {\color{purple}replacing} one word or phrase in the caption; or {\color{orange}swapping} two words or phrases within the caption. Although other forms of perturbations exist, we choose these two as they are the most well-represented in prior benchmarks.

\begin{figure*}[tb]
  \centering
  \includegraphics[width=0.74\linewidth]{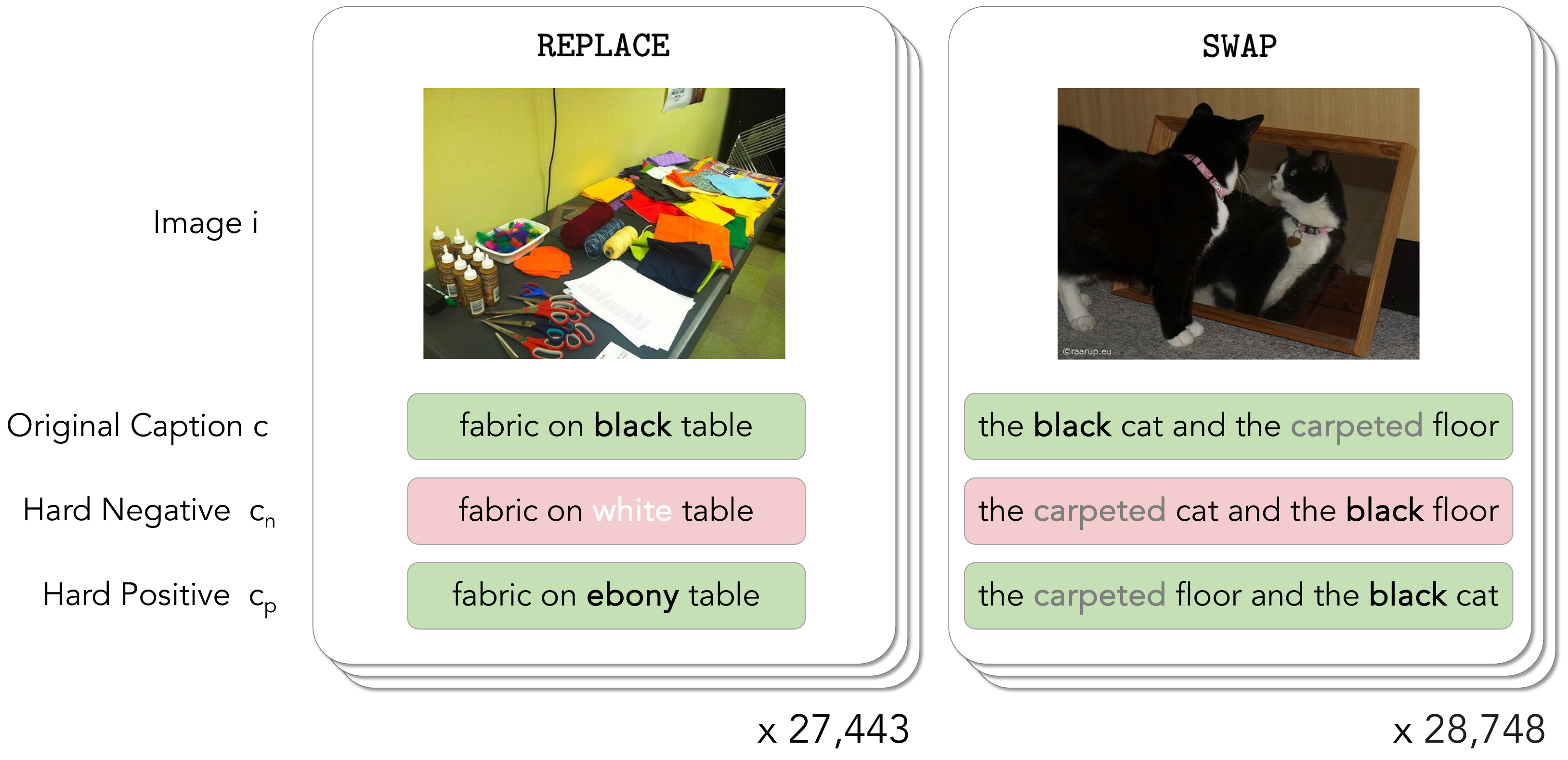}
  \caption{Our \replace and \swap evaluation sets. \replace replaces either an attribute or a relation in the original caption $c$ to obtain $c_n$ and $c_p$. \swap swaps object-attribute associations in the original caption $c$ to obtain $c_n$ and $c_p$.}
  \label{fig:benchmarks}
\end{figure*}

Therefore, we can consider each image in our dataset to be associated with three captions: the original caption $c$, a hard negative $c_n$ (sourced from an existing hard negative benchmark) and a hard positive $c_p$ (generated by us).
Figure~\ref{fig:benchmarks} shows examples from our benchmarks. \\

\noindent\textbf{Generating {\color{purple}replacements}.}
The most popular type of hard negative considered by existing work is \replace, where one word or phrase in the caption is replaced with another in a way that changes the meaning of the caption \cite{zhao2022vl, parcalabescu-etal-2022-valse, ma2022crepe, doveh2023dense, doveh2023teaching, kamath2023text, kamath-etal-2023-whats, hendricks-nematzadeh-2021-probing}.
To create hard positives, we replace one word or phrase in a way that does \textit{not} change the meaning of the caption. 

We begin with examples from VL-Checklist \cite{zhao2022vl}. This benchmark contains \replace hard negatives targeting either objects, attributes or relations. We focus on attributes and relations, as they have been shown to be more challenging for vision-language models to understand \cite{doveh2023dense, doveh2023teaching, hsieh2023sugarcrepe}, and select the subset of VL-Checklist based on Visual Genome \cite{krishna2017visual} to stay consistent with our \swap benchmark.
The VL-Checklist Relations benchmark has two types of relations: actions and spatial. The VL-Checklist Attributes benchmark has five types of attributes: action\footnote[1]{The action \textit{relation} is a transitive verb, e.g., ``a person wearing a shirt'', whereas the action \textit{attribute} is an intransitive verb, e.g., ``a person standing''.}, color, material, size, and state. 

For each of these types, we collect the ten most common relations/attributes, and hand-write a fixed replacement that holds for the various word senses of each original word. If no replacement can be found, we discard the sample. 
Finally, we replace 14 relations and 24 attributes, resulting in a benchmark of 16,868 hard positives targeting relations, and 10,575 hard positives targeting attributes, for a total of 27,443 examples (details in Appendix \ref{sec:additional_benchmark_details}).

E.g., for the Visual Genome caption ``cutting board next to pan'', VL-Checklist constructs a hard negative by replacing the relation with an antonym: ``cutting board \textit{far from} pan''. We construct a hard positive by replacing the relation with a synonym: ``cutting board \textit{near} pan''. 
While there may be minor differences between the original and hard positive captions (e.g., ``next to'' may imply a closer spatial relation than ``near''), they are both a match for the image, while the hard negative caption is not.

\noindent\textbf{Generating {\color{orange}swaps}.}
The other popular type of hard negative considered by existing work is \swap, where two words or phrases in a caption are swapped with each other in a way that changes the meaning of the caption \cite{yuksekgonul2023when, parcalabescu-etal-2022-valse, ma2022crepe, thrush2022winoground}. 
To create hard positives, we swap two phrases in a way that does \textit{not} change the meaning of the caption.

We begin with the Visual Genome Attribution (VGA) set from the Attribute-Relation-Order benchmark \cite{yuksekgonul2023when}, which switches object-attribute associations in a Visual Genome caption to create a hard negative, e.g., ``the crouched cat and the open door'' $\xrightarrow{}$ ``the open cat and the crouched door''. 
To create a hard positive, we switch the word order while retaining the object-attribute associations, thus retaining the meaning of the caption, e.g., ``the open door and the crouched cat''.
While there are small linguistic differences between the original and hard positive captions (e.g., people tend to describe the most salient object first), they are both a match for the image, where the hard negative caption is not. 
We create a hard positive for each example in the VGA dataset, resulting in a benchmark of 28,748 examples.


\section{Hard negative finetuning induces brittleness}
\label{sec:existing_models}


In this section we investigate existing models' performance, utilizing our more complete evaluation. We especially focus on evaluating whether recently introduced methods that train models with hard negatives indeed improve compositionality. \\



The goal of hard negative finetuning is to encourage CLIP models to understand how structural changes in language can affect the semantic interpretation of the caption. For example, finetuning on hard negatives targeting swaps should, in intuition, teach models that the directionality of a relationship between objects matters; finetuning on hard negatives targeting replacement should teach models to be sensitive to changes to any single word in the caption.
Ideally, we want the model to understand that perturbations to the caption (e.g., swaps, replacements) are important, and to recognize when a perturbed sentence has the same meaning as the original sentence, and when it does not.
However, we posit that solely emphasizing on hard negatives does not teach the model \textit{when} perturbations to the caption change meaning, they teach the model that perturbations \textit{do} change meaning, \textit{always}.

To validate our hypothesis, we benchmark a suite of CLIP models, trained regularly or with different hard negative augmentation strategies in Section~\ref{sec:eval_existing_models}.
We uncover that hard negative finetuning improves performance on hard negative evaluations at the cost of performance degradation on hard positives in Section~\ref{sec:results}. We finally discuss why this happens in Section~\ref{sec:why-this-happen}.

\setlength{\tabcolsep}{5pt}
\begin{table*}[t]
\centering
\resizebox{0.9\textwidth}{!}{  

            \begin{tabular}{llcccccc}
            \toprule
               && \multicolumn{2}{c}{\texttt{REPLACE}} & \multicolumn{2}{c}{\texttt{SWAP}} & \texttt{REPLACE} & \texttt{SWAP}\\
              &Model & \begin{tabular}{@{}c@{}}Orig. \\ Test Acc.\end{tabular} & \begin{tabular}{@{}c@{}}Aug. \\ Test Acc.\end{tabular} & \begin{tabular}{@{}c@{}}Orig. \\ Test Acc.\end{tabular}& \begin{tabular}{@{}c@{}}Aug. \\ Test Acc.\end{tabular}& Brittleness $(\downarrow)$& Brittleness$(\downarrow)$\\
              \midrule
              \textcolor{gray}{(a)}&CLIP ViT-B/32& 61.6 & 46.8 {\scriptsize\textcolor{red}{(-14.9)}}& 60.5 & 49.6 {\scriptsize\textcolor{red}{(-10.9)}} & 23.2 & 21.7\\
              \midrule
              &NegCLIP & \cellcolor{myblue}68.6 & \cellcolor{myblue}52.1 {\scriptsize\textcolor{red}{(-16.6)}}& \cellcolor{mypurple}70.9 & \cellcolor{mypurple}56.7 {\scriptsize\textcolor{red}{(-14.2)}}& \cellcolor{myblue}21.5 & \cellcolor{mypurple}26.4 \\
              &CREPE-Swap & \cellcolor{myblue}63.5 & \cellcolor{myblue}50.4 {\scriptsize\textcolor{red}{(-13.1)}}& \cellcolor{mypurple}70.6 & \cellcolor{mypurple}56.7 {\scriptsize\textcolor{red}{(-13.9)}}& \cellcolor{myblue}\textbf{19.8} & \cellcolor{mypurple}26.0\\
              &CREPE-Replace \hspace{20px}& \cellcolor{mypurple}73.7 & \cellcolor{mypurple}53.9 {\scriptsize\textcolor{red}{(-19.8)}}& \cellcolor{myblue}71.1 & \cellcolor{myblue}57.7 {\scriptsize\textcolor{red}{(-13.4)}}& \cellcolor{mypurple}23.9 & \cellcolor{myblue}25.4\\
              \textcolor{gray}{(b)}&SVLC & \cellcolor{mypurple}76.6 & \cellcolor{mypurple}44.5 {\scriptsize\textcolor{red}{(-32.1)}}& \cellcolor{myblue}72.4 & \cellcolor{myblue}\textbf{61.6} {\scriptsize\textcolor{red}{(-10.9)}}& \cellcolor{mypurple}39.9 & \cellcolor{myblue}\textbf{20.8}\\
              &SVLC+Pos & \cellcolor{mypurple}64.3 & \cellcolor{mypurple}45.0 {\scriptsize\textcolor{red}{(-19.3)}}& \cellcolor{myblue}56.5 & \cellcolor{myblue}45.4 {\scriptsize\textcolor{red}{(-11.1)}} & \cellcolor{mypurple}29.8 & \cellcolor{myblue}22.8\\
              &DAC-LLM & \cellcolor{mypurple}87.6 & \cellcolor{mypurple}48.9 {\scriptsize\textcolor{red}{(-38.7)}} & \cellcolor{myblue}72.0 & \cellcolor{myblue}61.1 {\scriptsize\textcolor{red}{(-10.9)}} & \cellcolor{mypurple}40.1 & \cellcolor{myblue}21.6 \\
              &DAC-SAM & \cellcolor{mypurple}86.9 & \cellcolor{mypurple}\textbf{55.9} {\scriptsize\textcolor{red}{(-31.0)}}& \cellcolor{myblue}69.5 & \cellcolor{myblue}56.5 {\scriptsize\textcolor{red}{(-13.0)}} & \cellcolor{mypurple}32.5 & \cellcolor{myblue}25.6 \\
              \midrule
              &Our HN & \cellcolor{mypurple}73.9 & \cellcolor{mypurple}55.7 {\scriptsize\textcolor{red}{(-18.2)}}& \cellcolor{mypurple}74.3 & \cellcolor{mypurple}60.5 {\scriptsize\textcolor{red}{(-13.8)}}& \cellcolor{mypurple}21.0 & \cellcolor{mypurple}25.1 \\
              \textcolor{gray}{(c)} &Our HP+HN & \cellcolor{mypurple}69.0 & \cellcolor{mypurple}\textbf{58.0} {\scriptsize\textcolor{red}{(-11.0)}}& \cellcolor{mypurple}73.2 & \cellcolor{mypurple}\textbf{61.1} {\scriptsize\textcolor{red}{(-12.1)}}& \cellcolor{mypurple}\textbf{16.9} & \cellcolor{mypurple}\textbf{22.9} \\
              \midrule
              &Our HP+HN (Swap-only) & \cellcolor{myblue}63.9 & \cellcolor{myblue}51.6 {\scriptsize\textcolor{red}{(-12.3)}}& \cellcolor{mypurple}73.0 & \cellcolor{mypurple}\textbf{61.9} {\scriptsize\textcolor{red}{(-11.2)}}& \cellcolor{myblue}18.6 & \cellcolor{mypurple}\textbf{21.2} \\
              \textcolor{gray}{(d)} &Our HP+HN (Replace-only) & \cellcolor{mypurple}70.9 & \cellcolor{mypurple}\textbf{59.0} {\scriptsize\textcolor{red}{(-11.9)}}& \cellcolor{myblue}69.7 & \cellcolor{myblue}55.6 {\scriptsize\textcolor{red}{(-14.1)}}& \cellcolor{mypurple}\textbf{17.8 }& \cellcolor{myblue}26.5\\
              \midrule
              &Random Chance & 50.0 & 33.3 & 50.0 & 33.3 & 33.3 & 33.3 \\
              &Human Estimate & 97 & 97 & 100 & 100 & 0 & 0\\
              \bottomrule \\
            \end{tabular}
        }
\caption{ Results of various ITM models on our benchmark: (a) CLIP; (b) Hard-Negative finetuned versions of CLIP from previous work (Section \ref{sec:results}); (c,d) Our improved model (Section \ref{sec:results_improved}). The purple cells indicate the models have seen perturbations of the type we are testing for during finetuning, blue cells indicate otherwise. \replace averages performance on Attributes and Relations; refer to Appendix \ref{sec:additional_results} for detailed results.}
\label{tab:results}
\end{table*}


\subsection{Evaluation}
\label{sec:eval_existing_models}

\noindent\textbf{Task.}
To evaluate model understanding of hard positives in addition to hard negatives, we use the image-text matching (ITM) task, consistent with existing benchmarks discussed in Section \ref{sec:protocol}. 
In our benchmark, the input is an image paired with three captions: two captions match the image (the original caption and the hard positive), and the third does not match the image (the hard negative). The model must return a high image-text matching score $s$ for the correct matching captions, and a low score for the incorrect one. \\

\noindent\textbf{Metrics.}
The first metric we use is the percentage of images in the benchmark for which the model-assigned score of the correct captions is higher than that of the incorrect caption.

For an image $i$, let the original caption be $c$, the hard negative from the existing benchmark (VGA for \swap and VL-Checklist for \replace) be $c_n$, and the hard positive that we construct (per Section \ref{sec:benchmarks}) be $c_p$. The vision-language model returns an image-text matching score $s(C|I)$ for some caption $C$ and image $I$. We measure \textit{Augmented Test Accuracy}: the fraction of instances in the benchmark where: 
\begin{equation}
\label{eq:1}
    s(c|i) > s(c_n|i) \text{ \hspace{0.1em} and \hspace{0.1em}  } s(c_p|i) > s(c_n|i)
\end{equation}

We do not require $s(c|i)$ to be equal to $s(c_p|i)$, as there are minor linguistic differences between the original caption and hard positive (c.f. Section \ref{sec:benchmarks}), and it is reasonable to predict that one of these captions matches the image slightly better than the other. However, as these two captions are both correct matches for the image and the hard negative is not, their model-assigned score should be higher than that of the hard negative caption.

The second metric we use is the percentage of images in the benchmark where the model treats $c$ and $c_p$ \textit{differently} when ranking with respect to $c_n$: ranking one of them above $c_n$ and one below. We measure this oversensitivity as \textit{Brittleness} $(\downarrow)$: the fraction of instances in the benchmark where:  
\begin{equation}
\label{eq:2}
\begin{split}
        s(c|i) > s(c_n|i) > s(c_p|i) \text{ \hspace{0.25em} or } \\ 
        s(c_p|i) > s(c_n|i) > s(c|i)
\end{split}
\end{equation} 

\vspace{1em}
\noindent\textbf{Random Chance Performance.} For Original Test Accuracy, random chance is 50\%, as there are only two possible rankings for the two captions (original and hard negative). For Augmented Test Accuracy, random chance is 33.3\%, as two of six possible rankings for the three captions (original, hard negative and hard positive) satisfy Condition (\ref{eq:1}).
For Brittleness, random chance is again 33.3\%, as two of six possible rankings for the three captions satisfy Condition (\ref{eq:2}). \\

\noindent\textbf{Human-estimated performance.}
We estimate human performance on our benchmark. We sample 100 data points each from the \swap and \replace benchmarks and solicit two expert annotations per data point. Each data point contains the image, the original caption, the hard negative and the hard positive. We ask the annotators to rank the captions based on the match for the image, allowing them to give multiple captions the same rank. 
The annotators have all taken at least one graduate-level course in NLP or Machine Learning. A point is awarded to the example if both annotators agree on the correct rank\footnote{
The errors in human performance on \replace arise from noise caused by errors in the underlying hard negative annotation (e.g., VL-Checklist containing a hard negative caption that is still a match for the image) or Visual Genome annotation (e.g., an incorrect region caption).}.
\newpage
\noindent\textbf{Models evaluated.}
Without loss of generality, we adopt the ViT-B/32 architecture for all our experiments. So, CLIP ViT-B/32 is our baseline CLIP model~\cite{RadfordKHRGASAM21}. We then evaluate several training interventions that finetune CLIP ViT-B/32 using different types of hard negatives:
NegCLIP~\cite{yuksekgonul2023when} is finetuned on hard negatives targeting word order shuffling; 
CREPE-Swap~\cite{ma2022crepe, hsieh2023sugarcrepe} is finetuned on hard negatives targeting single-phrase swaps; 
CREPE-Replace~\cite{ma2022crepe, hsieh2023sugarcrepe} is finetuned on hard negatives targeting single-phrase replacements; 
SVLC~\cite{doveh2023teaching} is finetuned on hard negatives targeting single-phrase replacements generated by LLMs and rule-based methods; 
SVLC+Pos~\cite{doveh2023teaching} is finetuned on the aforementioned hard negatives as well as paraphrases of the caption; 
DAC-LLM~\cite{doveh2023dense} is finetuned on several LLM-generated captions of the image as well as hard negatives generated by the SVLC method; 
and DAC-SAM~\cite{doveh2023dense} is finetuned on SAM-generated captions of the image as well as hard negatives generated by the SVLC method.

It is worth noting that SVLC+Pos, 
DAC-LLM and DAC-SAM 
contain ``positives'' in their finetuning, \ie, alternate captions that also match the image. However, these are not \textit{hard} positives, as in our work. Our alternate captions are \textit{minimal} perturbations to the original caption, swapping or replacing only single phrases while retaining the caption's meaning.

\subsection{Results}
\label{sec:results}
\noindent\textbf{Hard negative finetuning doesn't help models understand \textit{when} perturbations matter.}
In Table \ref{tab:results}, we first compare ITM model scores on only the original caption $c$ and the hard negative $c_n$, given an image $i$ --- as is done in existing work (Original Test Score). We then introduce the hard positive $c_p$ central to our work, and check: is the model score for the hard positive caption greater than that of the hard negative caption? Per Section \ref{sec:eval_existing_models}, we evaluate the cases when $s(c|i) > s(c_n|i) \text{ and } s(c_p|i) > s(c_n|i)$ (Augmented Test Score). 

We find that, when including hard positives, the performance of models finetuned on hard negatives drops (Aug. Test Score $<$ Orig. Test Score, difference depicted in red) by an average of 24.4 points for \replace and 12.5 points for \swap --- greater than the base model CLIP's 14.9 point and 10.9 point drops respectively. In fact, we see that as much as 39 points of model performance on hard negative benchmarks is misleading, as the model did not understand the underlying concept (e.g., word order) enough to recognize when the perturbation retained caption semantics.

\noindent\textbf{Hard negative finetuned models are oversensitive.}
Per Section \ref{sec:eval_existing_models}, to evaluate model brittleness, we calculate the percentage of instances in the benchmark where $s(c|i) > s(c_n|i) > s(c_p|i)$ or $s(c_p|i) > s(c_n|i) > s(c|i)$. In these instances, it is clear that the model does not understand that $c$ and $c_p$ have the same meaning and $c_n$ has a different meaning from both of them, \ie, it is oversensitive to the perturbation. In Table \ref{tab:results}, we see that in almost all cases, Brittleness increases after finetuning (rows (a) vs (b)) --- \ie, that hard negative finetuning makes the models more oversensitive to perturbations.

\noindent\textbf{Oversensitivity transfers across pertubation types.}
We see that, for each type of hard positive (\swap, \replace), the most oversensitive models are those finetuned on the corresponding hard negative (the purple cells in Table \ref{tab:results}), e.g., NegCLIP and CREPE-SWAP are finetuned on \swap hard negatives, and are the most oversensitive models under the \swap hard positives, and similarly for the other models on \replace. 
This is unsurprising, as the finetuning has taught the model to be sensitive to that specific type of perturbation.

However, we see that models trained on \replace hard negatives are still brittle to \swap hard positives (with an average score of 23.2), more so than the original CLIP baseline. We also see that models trained on \swap hard negatives are brittle to \replace hard positives (with an average score of 20.7), although less so than the original CLIP baseline --- potentially because a swap can be seen as two replacements. In essence, we see that the oversensitivity introduced by finetuning on hard negatives of one type of perturbation transfer to the other type of perturbation (blue cells in Table \ref{tab:results}).  

\noindent\textbf{``Non-hard'' positive finetuning increases oversensitivity.}
Three of the models we evaluate include finetuning on multiple correct captions (``positives'') for the image. For SVLC+Pos and DAC-LLM, these are generated by LLMs that see the caption alone, and for DAC-SAM, these are generated by BLIP2 \cite{li2023blip2} which sees segments of the image extracted by SAM \cite{kirillov2023segany}. 

However, c.f. Table \ref{tab:results}, this addition of positives to training does not improve model understanding of \textit{hard} positives compared to models finetuned on hard negatives alone; in fact, these models usually perform much worse. Comparing SVLC with SVLC+Pos, where the only difference is the addition of positives to training, it is clear that positive finetuning significantly increases oversensitivity.


Why? The alternate captions tend to be structurally very different from the original caption, and in the case of SAM-generated captions, contain different focuses entirely, as they only describe a segment of the image. Thus, they may give the model a more holistic understanding of the overall image \cite{doveh2023dense},
but not the fine-grained understanding we evaluate with our hard positives.

\noindent\textbf{Hard Negative finetuning lowers scores of the original captions too.}
Image-text matching scores are used to filter out data during web-scale corpora curation \cite{laion5b, gadre2023datacomp}, to evaluate captions for images \cite{hessel-etal-2021-clipscore}, to evaluate text-to-image generation \cite{saharia2022photorealistic, hu2023tifa}, and to evaluate text-to-video generation \cite{Ho2022ImagenVH}. Thus, while our evaluations focus on ranking, it is worth paying attention to the absolute value of the image text matching score itself. 

Across all benchmarks, models with hard negative finetuning lower the image-text matching score of the \textit{original} caption with the image as well --- not just the negative caption (c.f. Table \ref{tab:violin_plot} and Appendix \ref{sec:prevent_reduction}). 
In fact, the model that achieves one of the the highest performance on VL-Checklist, DAC-LLM, reduces the original caption scores on \replace from $0.23$ to $0.16$, a very large drop. 
This could cause significant errors in the aforementioned downstream applications. 
Examples are shown in Section \ref{sec:qual_analysis}.

\noindent\textbf{Different variants of CLIP all perform poorly.} In Appendix \ref{sec:additional_results}, we study the performance of CLIP with different model sizes, text encoders, pre-training data, and vision encoders. However, none of these variants significantly improve CLIP's poor compositionality on our benchmarks.

\subsection{Why does hard negative finetuning induce brittleness?}
\label{sec:why-this-happen}

From these results, it is clear that hard negative finetuning does not improve vision-language models' compositionality holistically. Performance on hard negatives is necessary but insufficient for compositionality, and by focusing on hard negatives alone, hard negative finetuning exacerbates poor performance on hard positives. We now discuss why the hard negative finetuning setup leads to worse performance on hard positives, as shown by our evaluation.

Let there be a set $\mathbb{P}$ of all possible small perturbations to the caption.  
During training on original captions and hard negatives alone, all perturbations $\mathcal{P} \in \mathbb{P}$ to the caption $c$ seen by the model $\mathcal{M}$ change the label of the caption.
The loss always penalizes $\mathcal{M}$ if $\mathcal{P}(c)$ matches the image under $\mathcal{M}$, \ie, the model is taught to reduce $s(\mathcal{P}(c)|i)$ for all seen $\mathcal{P}$. 
Thus, it is consistent with the training data to identify whether a text input $c$ somewhat matches the image and comes from the original caption distribution $\mathcal{C}$, and award it a high score if so, and a low score if not, \ie, if the caption appears to have been perturbed. Essentially, it is sufficient for $\mathcal{M}$ to learn perturbation detection. 

We see empirical proof of this in two ways (c.f. Section \ref{sec:results}): firstly, we see that $\mathcal{M}$ awards low scores to all perturbed captions, whether the meaning of the caption has changed or not; secondly, we see that this behavior transfers across \textit{types} of perturbations --- a model trained with \swap hard negatives awards low scores to \replace hard negatives and hard positives, and vice versa. 
Thus, by only showing models that perturbations \textit{do} change the input, not \textit{when} they change the input, we fail to attain improved compositionality.



\section{Exploring hard positive finetuning}


After establishing that finetuning on hard negatives alone teaches models that perturbations always change meaning, which causes poor compositionality, we explore a more well-rounded finetuning technique, incorporating hard positives into finetuning to determine whether that improves compositionality.



\subsection{Method}
We first generate hard positives using LLAMA-2 70B-Chat \cite{Touvron2023Llama2O}. We prompt this text-only model to modify a given caption without changing the meaning, either with word replacements, or swaps (if the caption contains the word ``and''). The inputs we provide the model are COCO-train captions. Prompting and generation details are provided in Appendix \ref{sec:hp_training_data_generation}. 

We then add these hard positives to model finetuning. We finetune CLIP ViT-B/32 on COCO-train with hard positives, generated as discussed above, and hard negatives, generated by the CREPE \cite{ma2022crepe} process, as in SugarCrepe \cite{hsieh2023sugarcrepe}. One hard positive and one hard negative is generated for each of the 591,753 COCO-train captions, resulting in an overall train set of 1,775,259 examples. We release this data to support further research in compositionality.

The finetuning follows the procedure outlined in SVLC \cite{doveh2023teaching}. 
We separately finetune CLIP ViT-B/32 on COCO-train with hard negatives only, to serve as a direct comparison for how the inclusion of hard positives in finetuning impacts model performance. We also finetune CLIP ViT-B/32 on COCO-train alone to serve as a control. Refer to Appendix \ref{sec:ft_implementation_details} for implementation details.

\subsection{Results}
\label{sec:results_improved}
\noindent\textbf{Adding hard positives to finetuning improves model performance.}
On \replace and \swap, our model finetuned on hard positives and hard negatives achieves the highest augmented test accuracy and lowest brittleness, compared to our model finetuned on hard negatives alone (Table \ref{tab:results}(c)). 

On \replace, our model also outperforms all hard negative finetuned models in Table \ref{tab:results}(b) in augmented test accuracy and brittleness. On \swap, our model outperforms NegCLIP, the CREPE-finetuned models, and DAC-SAM, but has slightly worse brittleness than the other models and slightly worse augmented test accuracy than SVLC. This could be due to the inherent difficulty of the \swap task --- not only could it be considered two replacements, but the word identities are unchanged, which causes added difficulty \cite{thrush2022winoground, yuksekgonul2023when}.  


\begin{table}[t]
\centering
\resizebox{0.8\columnwidth}{!}{  

            \begin{tabular}{lccc}
            \toprule
               & \multicolumn{3}{c}{Mean score} \\
              Model & \hspace{5px}$c$ $(\uparrow)$\hspace{5px} & $c_n$ $(\downarrow)$& $c_p$ $(\uparrow)$ \\
              \midrule
              CLIP ViT-B/32 & 0.234 & 0.226 & 0.229\\
              DAC-LLM & 0.160 & 0.134 & 0.131 \\
              Ours & 0.232 & 0.220 & 0.231\\
              
              \bottomrule \\
            \end{tabular}
        }
\caption{Mean score for $c$, $c_n$, and $c_p$ in \replace produced by CLIP, a hard negative finetuned model (DAC-LLM) and Our model. Our model exhibits better compositionality than CLIP and DAC-LLM by correctly lowering the score of $c_n$ but not $c$ or $c_p$. Refer to Appendix \ref{sec:prevent_reduction} for results across all models.
}
\label{tab:violin_plot}
\end{table}


\setlength{\tabcolsep}{5pt}
\begin{table*}[t]
\centering
\resizebox{0.9\textwidth}{!}{  

            \begin{tabular}{llcccccc}
            \toprule
               && \multicolumn{2}{c}{\texttt{REPLACE}} & \multicolumn{2}{c}{\texttt{SWAP}} & \texttt{REPLACE} & \texttt{SWAP}\\
              &Model & \begin{tabular}{@{}c@{}}Orig. \\ Test Acc.\end{tabular} & \begin{tabular}{@{}c@{}}Aug. \\ Test Acc.\end{tabular} & \begin{tabular}{@{}c@{}}Orig. \\ Test Acc.\end{tabular}& \begin{tabular}{@{}c@{}}Aug. \\ Test Acc.\end{tabular}& Brittleness $(\downarrow)$& Brittleness$(\downarrow)$\\
              \midrule
              \textcolor{gray}{(a)}&CLIP ViT-B/32& 61.6 & 46.8 {\scriptsize\textcolor{red}{(-14.9)}}& 60.5 & 49.6 {\scriptsize\textcolor{red}{(-10.9)}} & 23.2 & 21.7\\
              \midrule
              &$0$ HN & 58.5 & 49.8 {\scriptsize\textcolor{red}{(-8.6)}}& 64.1 & 51.2 {\scriptsize\textcolor{red}{(-12.9)}}& \textbf{15.8} & 25.0 \\
              &$0.25$ HN & 66.0 & 55.5 {\scriptsize\textcolor{red}{(-10.5)}}& 71.6 & 59.8 {\scriptsize\textcolor{red}{(-11.8)}}& 16.6 & 22.8 \\
              \textcolor{gray}{(b)} &$0.50$ HN & 67.3 & 56.9 {\scriptsize\textcolor{red}{(-10.5)}}& 72.5 & 60.5 {\scriptsize\textcolor{red}{(-12.0)}}& 16.4 & 22.8 \\
              &$0.75$ HN & 68.2 & \textbf{57.6} {\scriptsize\textcolor{red}{(-10.6)}}& 72.9 & \textbf{61.0} {\scriptsize\textcolor{red}{(-11.9)}}& 16.6 & \textbf{22.7} \\
              \midrule
              &Our HN & 73.9 & 55.7 {\scriptsize\textcolor{red}{(-18.2)}}& 74.3 & 60.5 {\scriptsize\textcolor{red}{(-13.8)}}& 21.0 & 25.1 \\
              \textcolor{gray}{(c)} &Our HP+HN & 69.0 & \textbf{58.0} {\scriptsize\textcolor{red}{(-11.0)}}& 73.2 & \textbf{61.1} {\scriptsize\textcolor{red}{(-12.1)}}& \textbf{16.9} & \textbf{22.9} \\
              \midrule
              &Random Chance & 50.0 & 33.3 & 50.0 & 33.3 & 33.3 & 33.3 \\
              &Human Estimate & 97 & 97 & 100 & 100 & 0 & 0\\
              \bottomrule \\
            \end{tabular}
        }
\caption{Results of ITM models on our benchmark while varying the ratio of hard negatives to hard positives during finetuning: (a) CLIP, (b) Ablated versions of our improved model, (c) Our improved model (Section \ref{sec:results_improved}). \replace averages performance on Attributes and Relations.}
\label{tab:hn_ablations}
\end{table*}


Table \ref{tab:violin_plot} shows the mean image-text matching scores of CLIP, DAC-LLM, and our finetuned model for the original, hard negative, and hard positive captions in \replace. CLIP awards similar scores to all, seeming to ignore the replacement for both hard negatives and hard positives. For DAC-LLM, the model recognizes the replacement for hard negatives and lowers the score significantly --- however, it incorrectly lowers the score of the hard positives by an even greater amount, although the meaning of the caption has not changed. Our finetuned model exhibits the correct behavior --- it reduces the score of the hard negative but maintains the score of the hard positive compared to the original caption. Moreover, unlike DAC-LLM, it does not lower the score of all captions, which could otherwise have repercussions downstream (c.f. Section \ref{sec:results}).

\noindent\textbf{Oversensitivity transfers across perturbations, but improved invariance does not.}
We additionally finetuned two CLIP ViT-B/32 models on hard positives and hard negatives targeting only \swap and only \replace respectively (c.f. Table \ref{tab:results}(d)). While neither of these models perform significantly better than the multi-task version on their respective evaluations (purple cells), we see that the Swap-Only finetuned model performs poorly on \replace, and likewise for the Replace-only finetuned model on \swap (blue cells). As such, while we saw that oversensitivity transferred across types of perturbations (Section \ref{sec:results}), it appears that improved invariance to a certain type of perturbation does not.

\noindent\textbf{Performance on standard benchmarks.}
In order to ensure that models do not experience catastrophic forgetting while finetuning on our data, we evaluate our finetuned models on standard benchmarks. As in \cite{yuksekgonul2023when}, we evaluate on ImageNet-1K, CIFAR-10, CIFAR-100, COCO Retrieval and Flickr30K Retrieval. 
Our models improve at hard positives and hard negatives while not losing overall performance. Refer to Appendix \ref{sec:standard_evals} for further details.



\subsection{Changing the ratio between hard positives and hard negatives}
In this section, we study the impact of changing the ratio between hard positive and hard negative losses during model finetuning. Table \ref{tab:hn_ablations} contains results of models trained on differing weights of hard negative loss while keeping the weight of hard positives loss fixed. 
We vary the weight of hard negative loss from $0$ (which equates to a model trained only on hard positives) to $1$ (which equates to our default proposed model, c.f. Table~\ref{tab:results}) in increments of $0.25$.

\begin{figure*}[tb]
  \centering
  \includegraphics[width=0.9\linewidth]{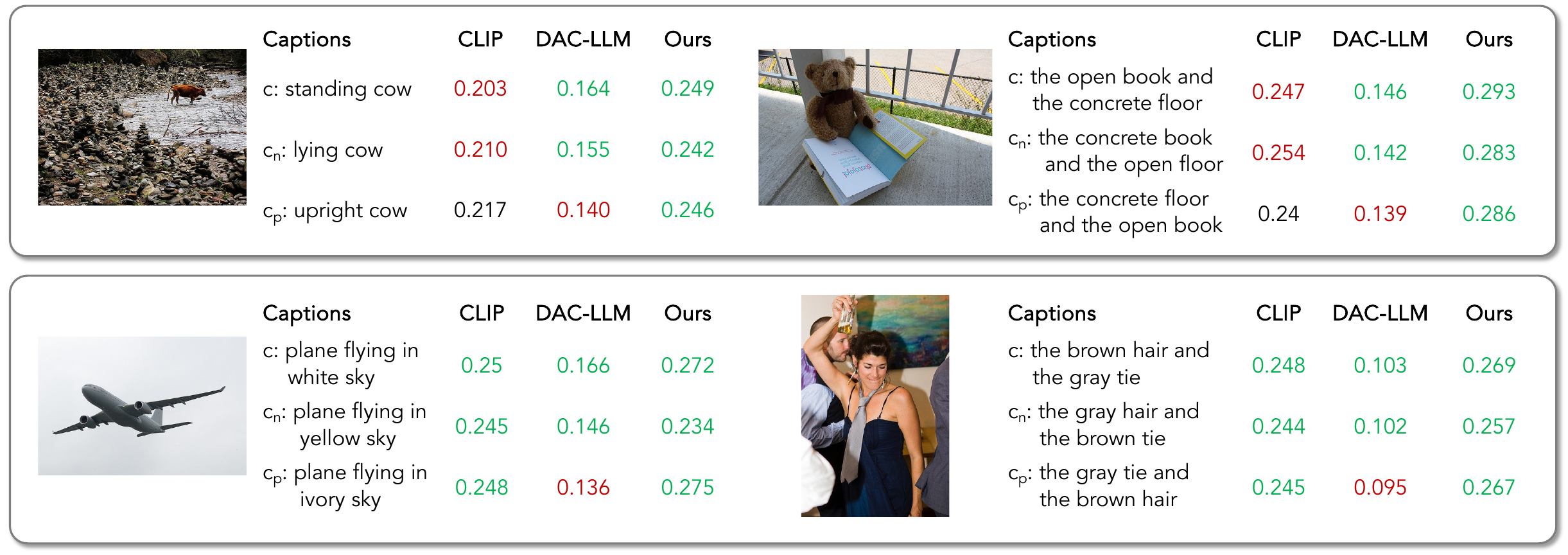}
  \caption{Sample predictions of CLIP, a hard negative finetuned model \cite{doveh2023dense}, and our model. Top: Considering hard negatives alone provides an incomplete picture of compositionality. Bottom: Hard negative finetuning can harm model performance. Both: Hard negative finetuning incorrectly lowers scores of the \textit{original} caption, unlike our model.}
  \label{fig:qual_analysis}
\end{figure*}

\noindent\textbf{Hard negatives are needed.} Rather unsurprisingly, the hard positive-only trained model performs poorly on our evaluation --- it has no sense of the existence of hard negatives, and learns from finetuning the \textit{opposite} of what hard negative-only finetuned models learn in existing work: rather than that perturbations \textit{always} change the label, this model learns that perturbations \textit{never} change the label. It is clear from these results that hard negatives are needed in addition to hard positives to improve model compositionality. 

\noindent\textbf{As the ratio of hard negatives to hard positives increases, test accuracy increases, but so may brittleness.} As the hard negative loss weight increases from $0$ to $1$, we see the Original and Augmented Test Accuracies both increasing. However, so too does the brittleness, for \replace. This trend continues: when the hard positives are dropped (i.e. a ratio of $\infty$), we see in Table \ref{tab:hn_ablations}(c) that the hard negative-only finetuned model achieves the highest Original Test Accuracy, but also has the highest brittleness for both \replace and \swap. This tradeoff suggests the need for careful tuning to achieve the best understanding of both hard positives and hard negatives.

\subsection{Qualitative Analysis}
\label{sec:qual_analysis}
Figure \ref{fig:qual_analysis} depicts examples of outputs of the original CLIP ViT-B/32 model, the hard-negative finetuned DAC-LLM, and our model finetuned on both hard positives and hard negatives. 

The top part shows similar behavior as depicted in Figure \ref{fig:teaser}: the hard negative finetuned model appears to have achieved high compositionality when its performance on $c$ and $c_n$ is compared to CLIP --- however, this is an incomplete picture. The hard negative finetuned model actually awards a lower score to $c_p$ than to $c_n$, showing that its understanding of compositionality is still lacking. In contrast, our model correctly awards higher scores to $c$ and $c_p$ than to $c_n$. 

The lower part shows instances of interesting behavior: where CLIP ranked the three captions correctly, and hard negative finetuning causes the model to now rank the captions incorrectly (awarding a low score to $c_P$). Clearly, hard negative finetuning can hurt the original model's performance.

In all shown examples, the hard negative finetuned model awards a lower score to \textit{all} captions than CLIP (including the \textit{original} caption), as discussed in Section \ref{sec:results}. Our model does not exhibit this behavior (c.f. Table \ref{tab:violin_plot} and Appendix \ref{sec:prevent_reduction}). 


\section{Discussion}


Our investigations explore a component of compositionality that has, until now, been largely underexplored. While a few efforts have studied the effects of training with positive rewritings~\cite{fan2023improving}, the use of \textit{hard} positives has been absent from the literature. We uncovered not just that CLIP models finetuned with hard negatives become oversensitive to changes, but that the de~facto CLIP model itself performs poorly on our augmented set. This calls into question whether CLIP models have a grounded sense of relational semantics~\cite{hsieh2023sugarcrepe}: for example, even basic text encoders such as word2vec~\cite{mikolov2013efficient, mikolov2013distributed} understand that ``white'' and ``ivory'' have closer meanings to each other than either does to ``blue'' --- so why should CLIP models fail to understand this, given \textit{additional} signal from the image, and millions of image-text pairs of supervision? 


Although training with hard positives mitigates the oversensitivity of CLIP models, models' performance is still far behind human performance. There is a need for further designs that incentivize compositionality by exploring alternative architecture designs and training objectives~\cite{bugliarello2023measuring, zeng2021multi, tschannen2024image}.
Our work calls for further research investigating more rigorously how finetuning methods targeting specific behaviors can cause adverse effects to overall model behavior, compared to the current status quo of simply evaluating on standard downstream evaluations. More research is also required to arrive at finetuning techniques that do not cause such adverse effects, and achieve the goal of improved robust vision-language compositionality.

\section*{Limitations} 
While we have further analysis in the Appendix, our work, like most work in vision-language compositionality today, is limited to CLIP-style models. There is a need to evaluate vision-language generation models, including Flamingo~\cite{alayrac2022flamingo}, BLIP~\cite{li2022blip, li2023blip2}, and GPT-4V~\cite{gpt-4v}, to isolate the effects of architecture and training objective. 
Additionally, while our models achieve higher performance on hard positives, more research is required to further improve performance and generalize to types of hard positives unseen during finetuning. 

\section*{Acknowledgements}
We thank Jieyu Zhang, Zixian Ma, the rest of the members of the RAIVN lab, William Merrill, as well as the anonymous reviewers, for helpful discussion and feedback. This work was partially supported by the Allen Institute for AI and ONR award N00014-23-1-2780.

\bibliographystyle{acl_natbib}
\bibliography{main}

\clearpage

\appendix

\section{Additional Benchmark Details}
\label{sec:additional_benchmark_details}
This section contains further details about the creation of the \replace benchmark, as well as a random sample of both benchmarks.

\subsection{Further details about \replace}
This dataset consists of hard negatives selected from VL-Checklist \cite{zhao2022vl} where one word or phrase in the caption is replaced with another in a way that changes the meaning of the caption, and hard positives we create where we replace one word or phrase in the caption with another in a way that does \textit{not} change the meaning of the caption. As discussed in Section \ref{sec:benchmarks}, we focus on the VL-Checklist hard negatives that target relations and attributes, as they are more challenging for models to understand. Additionally, we ignore objects because their replacements in VL-Checklist are not very targeted to be similar to the original object (e.g., positive: ``train has wheels'', negative: ``stir fry''), as the object class from which the hard negatives are created (all objects) is much broader than the relation or attribute classes (e.g., spatial relations, colors). We thus focus on relations and attributes, which have much harder hard negatives. We select the Visual Genome \cite{krishna2017visual} subset of VL-Checklist to stay consistent with the \swap benchmark, which is sourced from the same dataset.

The VL-Checklist Relations benchmark has two types of relations: actions and spatial. The VL-Checklist Attributes benchmark has five types of relations: action, color, material, size, and state. 
As discussed in Section \ref{sec:benchmarks}, for each of these types, we collect the ten most common relations/attributes, and hand-write a fixed replacement that holds for the various word senses of each original word. If no replacement can be found, we discard the sample. 
Finally, we replace 14 relations and 24 attributes, resulting in a benchmark of 16,868 hard positives targeting relations, and 10,575 hard positives targeting attributes, for a total of 27,443 examples.

The replaced relations and attributes, their replacements, their frequency in the benchmark, and an example caption containing each is provided in Tables \ref{tab:benchmark_details_rel}, \ref{tab:benchmark_details_att_1} and \ref{tab:benchmark_details_att_2}.

\vspace{4em}

\begin{table}[t]
\centering
\resizebox{\columnwidth}{!}{  

            \begin{tabular}{llcl}
            \toprule
              Orig. Rel. & Replaced Rel. & \hspace{5px}Freq.\hspace{5px} & Example \\
              \midrule
              in & within & 6173 & \begin{tabular}{@{}l@{}}\texttt{O: } white horse in field \\ \texttt{HP:} white horse within field \\ \texttt{HN:} white horse out of field\end{tabular}\\
              \midrule
              behind & to the rear of & 1057 & \begin{tabular}{@{}l@{}}\texttt{O: } van behind truck \\ \texttt{HP:} van to the rear of truck \\ \texttt{HN:} van in front of truck\end{tabular}\\
              \midrule
              on top of & on & 683 & \begin{tabular}{@{}l@{}}\texttt{O: } dishes on top of table \\ \texttt{HP:} dishes on table \\ \texttt{HN:} dishes below table\end{tabular}\\
              \midrule
              near & next to & 657 & \begin{tabular}{@{}l@{}}\texttt{O: } deck near water \\ \texttt{HP:} deck next to water \\ \texttt{HN:} deck far from water\end{tabular}\\
              \midrule
              next to & near & 621 & \begin{tabular}{@{}l@{}}\texttt{O: } person next to train \\ \texttt{HP:} person near train \\ \texttt{HN:} person far from train\end{tabular}\\
              \midrule
              under & beneath & 467 & \begin{tabular}{@{}l@{}}\texttt{O: } street under animals \\ \texttt{HP:} street beneath animals \\ \texttt{HN:} street above animals\end{tabular}\\
              \midrule
              by & near & 394 & \begin{tabular}{@{}l@{}}\texttt{O: } road by building \\ \texttt{HP:} road near building \\ \texttt{HN:} road far from building\end{tabular}\\
              \midrule
              above & on top of & 298 & \begin{tabular}{@{}l@{}}\texttt{O: } cloud above hill \\ \texttt{HP:} cloud on top of hill \\ \texttt{HN:} cloud below hill\end{tabular}\\
              \midrule
              wearing, wears & in & 3976 & \begin{tabular}{@{}l@{}}\texttt{O: } man wearing shirt  \\ \texttt{HP:} man in shirt \\ \texttt{HN:} man hugging shirt\end{tabular}\\
              \midrule
              holding & grasping & 950 & \begin{tabular}{@{}l@{}}\texttt{O: } woman holding fork \\ \texttt{HP:} woman grasping fork \\ \texttt{HN:} woman helping fork\end{tabular}\\
              \midrule
              sitting & seated & 639 & \begin{tabular}{@{}l@{}}\texttt{O: } cow sitting next to man \\ \texttt{HP:} cow seated next to man \\ \texttt{HN:} cow chasing man\end{tabular}\\
              \midrule
              hanging & dangling & 382 & \begin{tabular}{@{}l@{}}\texttt{O: } banner hanging from building \\ \texttt{HP:} banner dangling from building \\ \texttt{HN:} banner driving building\end{tabular}\\
              \midrule
              walking & strolling & 288 & \begin{tabular}{@{}l@{}}\texttt{O: } man walking on beach \\ \texttt{HP:} man strolling on beach \\ \texttt{HN:} man enclosing beach\end{tabular}\\
              \midrule
              riding on & traveling on & 283 & \begin{tabular}{@{}l@{}}\texttt{O:} person riding motorcycle \\ \texttt{HP:} person traveling on motorcycle \\ \texttt{HN:} person herding motorcycle\end{tabular}\\
              \bottomrule \\
            \end{tabular}
        }
\caption{Benchmark details of \replace Relations, which consist of spatial relations and transitive actions. \texttt{O}, \texttt{HP} and \texttt{HN} denote the Original, Hard Positive and Hard Negative captions respectively, randomly sampled from each relation. 
}
\label{tab:benchmark_details_rel}
\end{table}



\begin{table}[t]
\centering
\resizebox{0.8\columnwidth}{!}{  

            \begin{tabular}{llcl}
            \toprule
              Orig. Att. & Replaced Att. & Freq. & Example \\
              \midrule
              standing & upright & 153 & \begin{tabular}{@{}l@{}}\texttt{O: } turned head of a standing person \\ \texttt{HP:} turned head of a upright person \\ \texttt{HN:} turned head of a sitting person\end{tabular}\\
              \midrule
              sitting & seated & 88 & \begin{tabular}{@{}l@{}}\texttt{O: } sitting man \\ \texttt{HP:} seated man \\ \texttt{HN:} crouching man\end{tabular}\\
              \midrule
              walking & strolling & 64 & \begin{tabular}{@{}l@{}}\texttt{O: } foot of walking man \\ \texttt{HP:} foot of strolling man \\ \texttt{HN:} foot of lying man\end{tabular}\\
              \midrule
              eating & ingesting & 41 & \begin{tabular}{@{}l@{}}\texttt{O: } eating woman \\ \texttt{HP:} ingesting woman \\ \texttt{HN:} driving woman\end{tabular}\\
              \midrule
              hanging & dangling & 29 & \begin{tabular}{@{}l@{}}\texttt{O: } hanging branch \\ \texttt{HP:} dangling branch \\ \texttt{HN:} looking up branch\end{tabular}\\
              \midrule
              looking & gazing & 27 & \begin{tabular}{@{}l@{}}\texttt{O: } looking elephant \\ \texttt{HP:} gazing elephant \\ \texttt{HN:} playing elephant\end{tabular}\\
              \midrule
              white & ivory & 2742 & \begin{tabular}{@{}l@{}}\texttt{O: } white toilet \\ \texttt{HP:} ivory toilet \\ \texttt{HN:} orange toilet\end{tabular}\\
              \midrule
              black & ebony & 1790 & \begin{tabular}{@{}l@{}}\texttt{O: } black socks \\ \texttt{HP:} ebony socks \\ \texttt{HN:} dark brown socks\end{tabular}\\
              \midrule
              blue & sapphire & 1253 & \begin{tabular}{@{}l@{}}\texttt{O: } lady wearing blue shirt \\ \texttt{HP:} lady wearing sapphire shirt \\ \texttt{HN:} lady wearing yellow shirt\end{tabular}\\
              \midrule
              brown & chestnut & 947 & \begin{tabular}{@{}l@{}}\texttt{O: } edge of brown beach \\ \texttt{HP:} edge of chestnut beach \\ \texttt{HN:} edge of purple beach\end{tabular}\\
              \midrule
              red & crimson & 827 & \begin{tabular}{@{}l@{}}\texttt{O: } red glove \\ \texttt{HP:} crimson glove \\ \texttt{HN:} blue glove\end{tabular}\\
              \midrule
              green & emerald & 755 & \begin{tabular}{@{}l@{}}\texttt{O: } cooler has green lid \\ \texttt{HP:} cooler has emerald lid \\ \texttt{HN:} cooler has dark blue lid\end{tabular}\\
              \midrule
              silver & metallic & 242 & \begin{tabular}{@{}l@{}}\texttt{O: } silver fork \\ \texttt{HP:} metallic fork \\ \texttt{HN:} light brown fork\end{tabular}\\
              \bottomrule \\
            \end{tabular}
        }
\caption{Benchmark details of \replace Attributes (Part I, split due to space constraints), which consist of intransitive actions and colors. \texttt{O}, \texttt{HP} and \texttt{HN} denote the Original, Hard Positive and Hard Negative captions respectively, randomly sampled from each attribute. 
}
\label{tab:benchmark_details_att_1}
\end{table}



\begin{table}[t]
\centering
\resizebox{0.8\columnwidth}{!}{  

            \begin{tabular}{llcl}
            \toprule
              Orig. Att. & Replaced Att. & Freq. & Example \\
              \midrule
              large & big & 571 & \begin{tabular}{@{}l@{}}\texttt{O: } tire on large truck \\ \texttt{HP:} tire on big truck \\ \texttt{HN:} tire on tiny truck\end{tabular}\\
              \midrule
              small & tiny & 358 & \begin{tabular}{@{}l@{}}\texttt{O: } toilet inside small bathroom \\ \texttt{HP:} toilet inside tiny bathroom \\ \texttt{HN:} toilet inside huge bathroom\end{tabular}\\
              \midrule
              long & lengthy & 271 & \begin{tabular}{@{}l@{}}\texttt{O: } person carrying a long skateboard \\ \texttt{HP:} person carrying a lengthy skateboard \\ \texttt{HN:} person carrying a short skateboard\end{tabular}\\
              \midrule
              big & large & 146 & \begin{tabular}{@{}l@{}}\texttt{O: } big elephant \\ \texttt{HP:} large elephant \\ \texttt{HN:} tiny elephant\end{tabular}\\
              \midrule
              huge & big & 31 & \begin{tabular}{@{}l@{}}\texttt{O: } kites under huge sky \\ \texttt{HP:} kites under big sky \\ \texttt{HN:} kites under tiny sky\end{tabular}\\
              \midrule
              wet & damp & 62 & \begin{tabular}{@{}l@{}}\texttt{O: } wet road \\ \texttt{HP:} damp road \\ \texttt{HN:} cloudless road \end{tabular}\\
              \midrule
              smiling & happy & 50 & \begin{tabular}{@{}l@{}}\texttt{O: } snowboard with smiling man \\ \texttt{HP:} snowboard with happy man \\ \texttt{HN:} snowboard with sad man\end{tabular}\\
              \midrule
              old & aged & 46 & \begin{tabular}{@{}l@{}}\texttt{O: } old train \\ \texttt{HP:} aged train \\ \texttt{HN:} young train\end{tabular}\\
              \midrule
              clear & unclouded & 43 & \begin{tabular}{@{}l@{}}\texttt{O: } clear sky \\ \texttt{HP:} unclouded sky \\ \texttt{HN:} partly cloudy sky\end{tabular}\\
              \midrule
              young & youthful & 36 & \begin{tabular}{@{}l@{}}\texttt{O: } shoes on young man \\ \texttt{HP:} shoes on youthful man \\ \texttt{HN:} shoes on unhappy man\end{tabular}\\
              \bottomrule \\
            \end{tabular}
        }
\caption{Benchmark details of \replace Attributes (Part II, split due to space constraints), which consist of sizes and states. The fifth attribute, material, had no synonyms for each word (e.g., ``brick''), so we discard it. \texttt{O}, \texttt{HP} and \texttt{HN} denote the Original, Hard Positive and Hard Negative captions respectively, randomly sampled from each attribute. 
}
\label{tab:benchmark_details_att_2}
\end{table}



\subsection{Random samples of \replace and \swap}
\begin{figure*}[tb]
  \centering
  \includegraphics[width=0.85\linewidth]{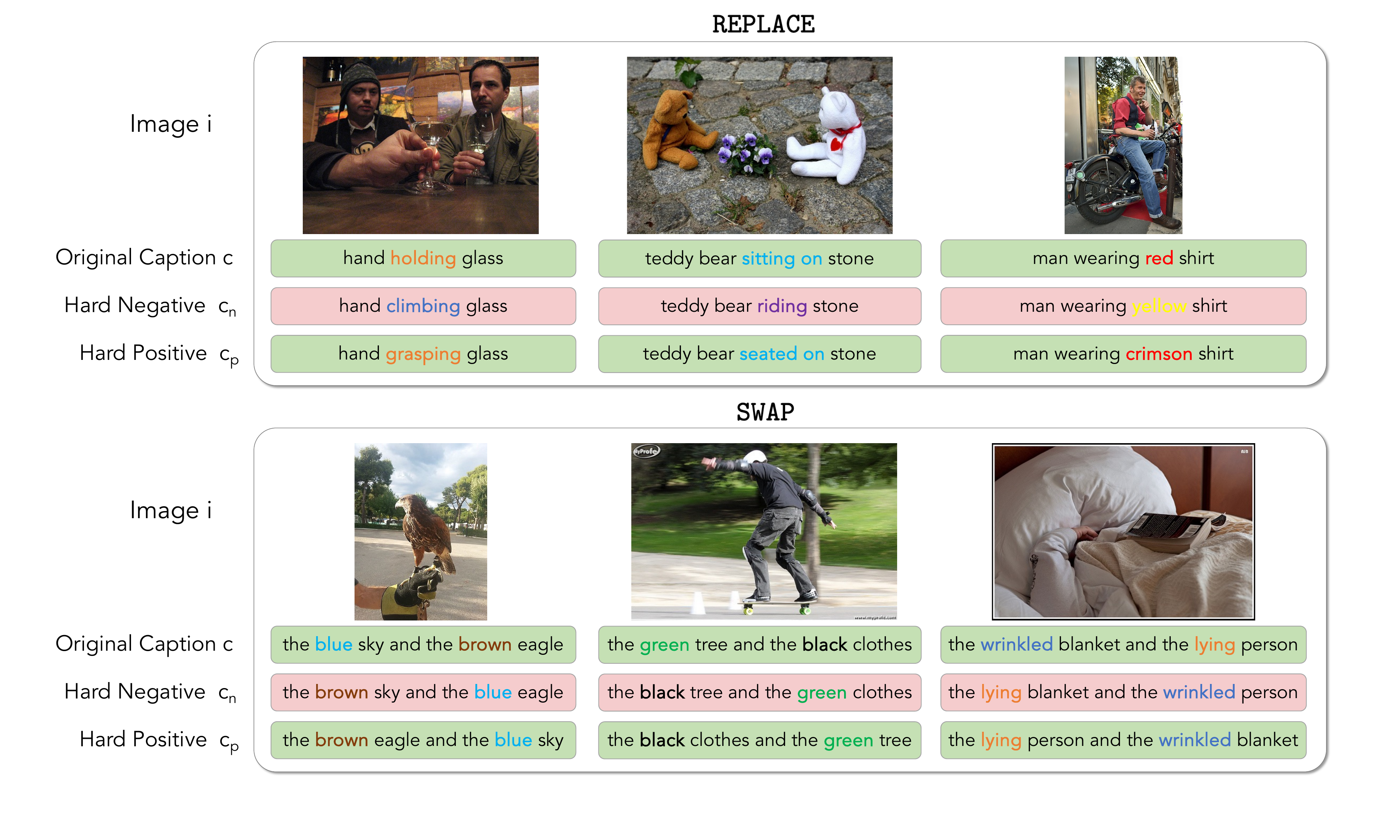}
  \caption{Random samples of \replace and \swap. The first two \replace samples are from Relations, and the third from Attributes.}
  \label{fig:random_sample}
\end{figure*}

Figure \ref{fig:random_sample} contains random samples of \replace-Relations, \replace-Attributes and \swap. As the benchmarks are created from Visual Genome region annotations, they occasionally only discuss a part of the image; however, the hard negative captions are created such that they are always a mismatch for the corresponding image --- i.e., they do not satisfy any part of the image \cite{zhao2022vl, yuksekgonul2023when}. 

\clearpage
\section{Additional Results}
\label{sec:additional_results}
\vspace{-8em}
This section contains additional results, splitting the \replace results in the main paper into the separate Relations and Attributes subsets (Table \ref{tab:replace_split}), as well as the results of various other models on our benchmarks: varying model size, architecture, pretraining data, and training objective (Table \ref{tab:more_models}). 




\vspace{-2.5em}

\noindent\textbf{Replacing relations vs replacing attributes.} Table \ref{tab:replace_split} contains the results for the models in the main paper, split across \replace Relations and \replace Attributes. It is clear that model performance is worse on Relations, likely because relations are more challenging than attributes for models to understand --- following simple combinatorial logic, it is more likely that within one training batch, the same \textit{object} appears twice with different attributes, than that the same \textit{pair of objects} appears twice with different relations between them. This contributes towards why contrastively trained models better understand attributes than relations. 

\vspace{-3em}

Following a similar trend, our model finetuned on both hard positives and hard negatives performs extremely well on \replace-Attributes (more so than on \replace-Relations), achieving high Augmented Test Accuracy and low Brittleness --- in fact, the drop from Original Accuracy is only 6.7 points, almost four times lower than the average drop of 24.7 points across models from existing work. 

\vspace{-2.5em}

\noindent\textbf{Changing CLIP model size.} From Table \ref{tab:more_models}(b), it is clear that increasing the model size of CLIP does not necessarily improve its performance on our benchmarks --- there is no clear pattern in the results of various models. 

\vspace{-3em}

\noindent\textbf{Changing CLIP text encoder.} From Table \ref{tab:more_models}(c), we see the effect of using pretrained RoBERTa weights in the CLIP text encoder. The model performance is fair for \replace, but very poor for \swap --- likely due to the fact that only the word order changes across all three captions, and masked language models have been shown to struggle with word order. 
\vspace{-3em}

\noindent\textbf{Changing CLIP pretraining data.} From Table \ref{tab:more_models}(d), using DataComp \cite{gadre2023datacomp} as the pretraining data for CLIP seems to hurt model performance, more so on \replace than on \swap. \\
\vspace{-4em}

\noindent\textbf{Changing CLIP vision encoder.} From Table \ref{tab:more_models}(e), replacing the ViT vision encoder with a ResNet-based vision encoder seems to improve performance slightly, in the case of RN50 models. \\

\noindent\textbf{Comparing CLIP to XVLM \cite{zeng22c}. } Table \ref{tab:more_models}(f) shows the performance of XVLM-16M (pretrained) on our benchmarks, as it has been shown to perform well on hard negative-focused benchmarks \cite{bugliarello2023measuring}. At first glance, the performance is shockingly high compared to CLIP --- however, it is important to note that XVLM is trained on Visual Genome region captions, from which all of our benchmarks are sourced. It is possible that there is data leakage, as the XVLM training data was curated to prevent leakage with popular test sets \textit{at the time}, and pre-dates ARO \cite{yuksekgonul2023when} and VL-Checklist \cite{zhao2022vl}, from which our benchmarks are sourced. This may also explain the results of \cite{bugliarello2023measuring}.

\setlength{\tabcolsep}{5pt}
\begin{table*}[t]
\centering
\resizebox{0.9\textwidth}{!}{  

            \begin{tabular}{llcccccc}
            \toprule
               && \multicolumn{2}{c}{\texttt{REPLACE-Rel}} & \multicolumn{2}{c}{\texttt{REPLACE-Att}} & \texttt{REPLACE-Rel} & \texttt{REPLACE-Att}\\
              &Model & \begin{tabular}{@{}c@{}}Orig. \\ Test Acc.\end{tabular} & \begin{tabular}{@{}c@{}}Aug. \\ Test Acc.\end{tabular} & \begin{tabular}{@{}c@{}}Orig. \\ Test Acc.\end{tabular}& \begin{tabular}{@{}c@{}}Aug. \\ Test Acc.\end{tabular}& Brittleness $(\downarrow)$& Brittleness$(\downarrow)$\\
              \midrule
              \textcolor{gray}{(a)}&CLIP ViT-B/32& 57.6 & 45.3 {\scriptsize\textcolor{red}{(-12.3)}}& 68.1 & 49.0 {\scriptsize\textcolor{red}{(-19.1)}} & 21.7 & 25.5\\
              \midrule
              &NegCLIP & \cellcolor{myblue}65.6 & \cellcolor{myblue}48.2 {\scriptsize\textcolor{red}{(-17.4)}}& \cellcolor{myblue}73.4 & \cellcolor{myblue}58.2 {\scriptsize\textcolor{red}{(-15.2)}}& \cellcolor{myblue}22.3 & \cellcolor{myblue}20.3 \\
              &CREPE-Swap & \cellcolor{myblue}56.6 & \cellcolor{myblue}43.0 {\scriptsize\textcolor{red}{(-13.7)}}& \cellcolor{myblue}74.4 & \cellcolor{myblue}\textbf{62.2} {\scriptsize\textcolor{red}{(-12.1)}}& \cellcolor{myblue}\textbf{21.2} & \cellcolor{myblue}17.6\\
              &CREPE-Replace \hspace{20px}& \cellcolor{mypurple}70.5 & \cellcolor{mypurple}49.4 {\scriptsize\textcolor{red}{(-21.1)}}& \cellcolor{mypurple}78.8 & \cellcolor{mypurple}61.1 {\scriptsize\textcolor{red}{(-17.7)}}& \cellcolor{mypurple}25.3 & \cellcolor{mypurple}21.6\\
              \textcolor{gray}{(b)}&SVLC & \cellcolor{mypurple}72.0 & \cellcolor{mypurple}42.1 {\scriptsize\textcolor{red}{(-29.9)}}& \cellcolor{mypurple}83.8 & \cellcolor{mypurple}48.2 {\scriptsize\textcolor{red}{(-35.6)}}& \cellcolor{mypurple}41.6 & \cellcolor{mypurple}\textbf{37.3}\\
              &SVLC+Pos & \cellcolor{mypurple}62.1 & \cellcolor{mypurple}44.7 {\scriptsize\textcolor{red}{(-17.4)}}& \cellcolor{mypurple}68.0 & \cellcolor{mypurple}45.6 {\scriptsize\textcolor{red}{(-22.4)}} & \cellcolor{mypurple}30.3 & \cellcolor{mypurple}29.0\\
              &DAC-LLM & \cellcolor{mypurple}88.1 & \cellcolor{mypurple}51.5 {\scriptsize\textcolor{red}{(-36.6)}} & \cellcolor{mypurple}86.8 & \cellcolor{mypurple}44.9 {\scriptsize\textcolor{red}{(-41.9)}} & \cellcolor{mypurple}38.4 & \cellcolor{mypurple}42.7 \\
              &DAC-SAM & \cellcolor{mypurple}89.2 & \cellcolor{mypurple}\textbf{59.6} {\scriptsize\textcolor{red}{(-29.5)}}& \cellcolor{mypurple}86.9 & \cellcolor{mypurple}55.9 {\scriptsize\textcolor{red}{(-31.0)}} & \cellcolor{mypurple}31.2 & \cellcolor{mypurple}32.5 \\
              \midrule
              &Our HN & \cellcolor{mypurple}71.6 & \cellcolor{mypurple}\textbf{52.6} {\scriptsize\textcolor{red}{(-19.0)}}& \cellcolor{mypurple}77.5 & \cellcolor{mypurple}60.8 {\scriptsize\textcolor{red}{(-16.8)}}& \cellcolor{mypurple}23.5 & \cellcolor{mypurple}21.0 \\
              \textcolor{gray}{(c)} &Our HP+HN & \cellcolor{mypurple}65.5 & \cellcolor{mypurple} 51.9 {\scriptsize\textcolor{red}{(-13.6)}}& \cellcolor{mypurple}74.5 & \cellcolor{mypurple}\textbf{67.7} {\scriptsize\textcolor{red}{(-6.7)}}& \cellcolor{mypurple}\textbf{19.9} & \cellcolor{mypurple}\textbf{12.2} \\
              \midrule
              &Our HP+HN (Swap-only) & \cellcolor{myblue}57.0 & \cellcolor{myblue}44.4 {\scriptsize\textcolor{red}{(-12.6)}}& \cellcolor{myblue}75.1 & \cellcolor{myblue}63.1 {\scriptsize\textcolor{red}{(-11.9)}}& \cellcolor{myblue}\textbf{19.4} & \cellcolor{myblue}17.2 \\
              \textcolor{gray}{(d)} &Our HP+HN (Replace-only) & \cellcolor{mypurple}68.8 & \cellcolor{mypurple}\textbf{53.7} {\scriptsize\textcolor{red}{(-15.1)}}& \cellcolor{mypurple}74.2 & \cellcolor{mypurple}\textbf{67.3} {\scriptsize\textcolor{red}{(-6.8)}}& \cellcolor{mypurple}21.0& \cellcolor{mypurple}\textbf{12.7}\\
              \midrule
              &Random Chance & 50.0 & 33.3 & 50.0 & 33.3 & 33.3 & 33.3 \\
              &Human Estimate & 97 & 97 & 100 & 100 & 0 & 0\\
              \bottomrule \\
            \end{tabular}
        }
\caption{ Detailed results of various ITM models on our \replace benchmark: (a) CLIP, (b) Hard-Negative finetuned versions of CLIP from previous work (Section \ref{sec:results}), (c) Our improved model (Section \ref{sec:results_improved}). The purple cells indicate the models have seen perturbations of the type we are testing for during finetuning, blue cells indicate otherwise. We report performance on the Relations and Attributes subsets of \replace separately here; they are averaged in the main paper for brevity.}
\label{tab:replace_split}
\end{table*}


\setlength{\tabcolsep}{5pt}
\begin{table*}[t]
\centering
\resizebox{0.9\textwidth}{!}{  

            \begin{tabular}{llcccccc}
            \toprule
               && \multicolumn{2}{c}{\texttt{REPLACE}} & \multicolumn{2}{c}{\texttt{SWAP}} & \texttt{REPLACE} & \texttt{SWAP}\\
              &Model & \begin{tabular}{@{}c@{}}Orig. \\ Test Acc.\end{tabular} & \begin{tabular}{@{}c@{}}Aug. \\ Test Acc.\end{tabular} & \begin{tabular}{@{}c@{}}Orig. \\ Test Acc.\end{tabular}& \begin{tabular}{@{}c@{}}Aug. \\ Test Acc.\end{tabular}& Brittleness $(\downarrow)$& Brittleness$(\downarrow)$\\
              \midrule
              \textcolor{gray}{(a)}&CLIP ViT-B/32& 61.6 & 46.8 {\scriptsize\textcolor{red}{(-14.9)}}& 60.5 & 49.6 {\scriptsize\textcolor{red}{(-10.9)}} & 23.2 & 21.7\\
              \midrule
              &CLIP ViT-B/16& 61.8 & 45.0 {\scriptsize\textcolor{red}{(-16.8)}}& 61.1 & 51.1 {\scriptsize\textcolor{red}{(-10.0)}} & 24.8 & 19.8\\
              &CLIP ViT-L/14& 64.2 & 48.4 {\scriptsize\textcolor{red}{(-15.8)}}& 61.1 & 49.9 {\scriptsize\textcolor{red}{(-11.2)}} & 24.0 & 21.9\\
              \textcolor{gray}{(b)}&OpenCLIP ViT-H/14& 56.5 & 43.7 {\scriptsize\textcolor{red}{(-12.8)}}& 62.9 & 51.7 {\scriptsize\textcolor{red}{(-11.2)}} & 20.5 & 21.7\\
              &OpenCLIP ViT-g/14& 59.5 & 45.8 {\scriptsize\textcolor{red}{(-13.7)}}& 63.5 & 52.1 {\scriptsize\textcolor{red}{(-11.4)}} & 22.2 & 22.4\\
              &OpenCLIP ViT-G/14& 58.6 & 44.4 {\scriptsize\textcolor{red}{(-14.2)}}& 61.9 & 50.5 {\scriptsize\textcolor{red}{(-11.3)}} & 22.9 & 22.4\\
              \midrule
              \textcolor{gray}{(c)}&RoBERTa-CLIP ViT-B/32& 57.5 & 44.3 {\scriptsize\textcolor{red}{(-13.3)}}& 48.7 & 29.4 {\scriptsize\textcolor{red}{(-19.3)}} & 28.7 & 40.3\\
              \midrule
              &DataComp-CLIP ViT-B/32& 53.0 & 42.4 {\scriptsize\textcolor{red}{(-10.6)}}& 58.5 & 44.8 {\scriptsize\textcolor{red}{(-13.7)}} & 21.2 & 27.1\\
              \textcolor{gray}{(d)}&DataComp-CLIP ViT-B/16& 51.7 & 40.8 {\scriptsize\textcolor{red}{(-10.9)}}& 56.8 & 43.6 {\scriptsize\textcolor{red}{(-13.2)}} & 21.5 & 26.5\\
              &DataComp-CLIP ViT-L/14& 55.7 & 42.7 {\scriptsize\textcolor{red}{(-13.1)}}& 60.0 & 47.6 {\scriptsize\textcolor{red}{(-12.4)}} & 22.0 & 24.2\\
              \midrule
              &CLIP-RN50x16& 63.2 & 45.8 {\scriptsize\textcolor{red}{(-17.5)}}& 62.2 & 51.9 {\scriptsize\textcolor{red}{(-10.3)}} & 24.9 & 20.0\\
              \textcolor{gray}{(e)}&CLIP-RN50x64& 66.3 & 49.2 {\scriptsize\textcolor{red}{(-17.1)}}& 62.2 & 51.3 {\scriptsize\textcolor{red}{(-10.9)}} & 25.4 & 21.1\\
              &CLIP-RN101& 58.3 & 43.9 {\scriptsize\textcolor{red}{(-14.4)}}& 61.9 & 52.0 {\scriptsize\textcolor{red}{(-9.9)}} & 23.2 & 19.3\\
              \midrule
              \textcolor{gray}{(f)}&XVLM-16M*& 72.9 & 63.8 {\scriptsize\textcolor{red}{(-9.1)}}& 89.3 & 84.8 {\scriptsize\textcolor{red}{(-4.5)}} & 16.3 & 8.1\\
              \midrule
              &Random Chance & 50.0 & 33.3 & 50.0 & 33.3 & 33.3 & 33.3 \\
              &Human Estimate & 97 & 97 & 100 & 100 & 0 & 0\\
              \bottomrule \\
            \end{tabular}
        }
\caption{ Results of additional ITM models on our benchmark: (a) CLIP, (b) Different model sizes of CLIP, (c) CLIP where the text encoder is initialized with RoBERTa-pretrained weights, (d) CLIP trained on DataComp \cite{gadre2023datacomp} rather than WIT \cite{RadfordKHRGASAM21} or LAION \cite{schuhmann2022laion}, (e) CLIP with different vision encoders, (f) XVLM*. The * on XVLM depicts that it is not a fair comparison with the other models, as XVLM is trained specifically on VG region captions, from which our benchmarks are sourced. \replace averages performance on Attributes and Relations.}
\label{tab:more_models}
\end{table*}


\section{Hard Positive Training Data Generation Details}
\label{sec:hp_training_data_generation}
In this section, we discuss the details of generating hard positive training data. First, we discuss the prompts used to generate data from the LLM LLAMA2 \cite{Touvron2023Llama2O}. Then, we discuss the implementation details of the generation. Finally, we provide a random sample of the data generated using the prompts. 


\subsection{Prompts}
The prompt for \replace is:\\

{\fontfamily{qcr}\selectfont \scriptsize
\noindent Replace one word in this sentence with a synonym, without changing the meaning of the sentence. Only output the changed sentence.

\noindent \{example\}
}  \\

\noindent The prompt for \swap is: \\

{\fontfamily{qcr}\selectfont \scriptsize
\noindent Swap the words around the word "and" in a sentence without changing the meaning. Only respond with the changed sentence. \\

\noindent Input: three giraffes and two antelope

\noindent Output: two antelopes and three giraffes

\noindent Input: a blue and white stained glass clock shows the time

\noindent Output: a white and blue stained glass clock shows the time 

\noindent Input: a mixture of rice and broccoli are put together

\noindent Output: a mixture of broccoli and rice are put together 

\noindent Input: a bathroom with a sink, toilet and shower

\noindent Output: a bathroom with a sink, shower and toilet 

\noindent Input: there is a man wearing glasses and holding a wine bottle

\noindent Output: there is a man holding a wine bottle and wearing glasses \\

\noindent Input: \{example\}

\noindent Output:
}

\clearpage
\noindent We arrived at the examples in the \swap prompt by looking at patterns of common mistakes in the LLM outputs. No such examples were needed for \replace, as it appears to be an easier task, e.g., not requiring correct dependency parsing of text inputs, which can be potentially ungrammatical captions. 

\vspace{-6em}

\subsection{Implementation details}
\vspace{-3em}
We generate hard positive training data by feeding the above prompt to the LLAMA2 70B-Chat model \cite{Touvron2023Llama2O}. The examples are sourced from COCO train (note: Hard negatives are generated from COCO train as well, following the CREPE \cite{ma2022crepe} procedure). \swap hard positives are created for COCO train captions containing the word ``and'' and less than 15 words, which amounts to $119,071$ captions, and \replace hard positives are created for all $591,753$ COCO train captions. In total, we generate $710,824$ hard positives --- although we subsample these during finetuning, as discussed in Section \ref{sec:ft_implementation_details}. 

\vspace{-1.5em}

We run inference on LLAMA2 with Flash Attention on a batch size of 32, on 4xA100s, which takes 36 hours to generate all hard positives (we parallelize this across 8 similar machines). For \swap we set the maximum number of generated tokens to 20 (as we filter out captions of greater than 15 words), and for \replace we set it to 30 (as we do no such filtering).

\vspace{-1em}

Note: We considered using Spacy to get dependency parses of the sentences and write code to perform the swapping, but Spacy fails often on COCO image captions, which are often only noun phrases (e.g., ``a person on a brown horse'') or ungrammatical. Thus, we used an LLM instead, which had almost perfect performance in swapping sentences from a random sample of 100 inputs we went through manually.

\vspace{-6em}
\subsection{Random sample of generated data}
\vspace{-3em}

Below is a random sample of the generated data for \swap:\\

\vspace{-1em}
{\fontfamily{qcr}\selectfont \scriptsize
\noindent A cabinet setting with green vases and a wooden backboard $\xrightarrow[]{}$ 

\vspace{-1em}
\noindent A cabinet setting with a wooden backboard and green vases

\noindent A couch and a television in a room $\xrightarrow[]{}$

\noindent A television and a couch in a room

\noindent An older gentleman in a white shirt and black bow tie $\xrightarrow[]{}$

\noindent An older gentleman in a black bow tie and white shirt \\

\noindent Two giraffes standing next to one another with trees and bushes near them $\xrightarrow[]{}$

\noindent Two giraffes standing next to one another with bushes and trees near them \\

\noindent a lady wearing snow skis and a man holding snow skis $\xrightarrow[]{}$

\noindent a man holding snow skis and a lady wearing snow skis \\

\noindent An adorable little girl wearing sunglasses and holding a stack of frisbee $\xrightarrow[]{}$

\noindent An adorable little girl holding a stack of frisbee and wearing sunglass  \\

}

\noindent Below is a random sample of the generated data for \replace: \\

{\fontfamily{qcr}\selectfont \scriptsize
\noindent a person holding an piece of an eaten sandhwich next to a lap top computer $\xrightarrow[]{}$ 

\noindent a person holding a morsel of a devoured sandwich next to a portable computer \\

\noindent Two baby goats stand together on worn stones $\xrightarrow[]{}$

\noindent Two baby kids stand together on worn rocks \\

\noindent a field that ha a bunch of sheep in it $\xrightarrow[]{}$

\noindent a meadow that has a flock of sheep in it \\

\noindent A side view mirror on the handle bars of a motorcycle $\xrightarrow[]{}$

\noindent A side view mirror on the handle bars of a motorbike \\

\noindent A variety of vegetables sits in a pile on a stand $\xrightarrow[]{}$

\noindent A collection of vegetables sits in a pile on a stand \\

\noindent a man going down a handle on some stairs on a skate board $\xrightarrow[]{}$

\noindent a man going down a rail on some stairs on a skate board \\

}

We notice that the LLM frequently changes grammatical errors if present in the original caption when generating the hard positive caption, e.g., ``a field that \textit{ha} ...'' $\xrightarrow[]{}$ ``a meadow that \textit{has} ...''. 

We also notice that, while generating \replace hard positives, the LLM tends to replace the objects (``field'' $\xrightarrow[]{}$ ``meadow''), more than the attributes (``eaten'' $\xrightarrow[]{}$ ``devoured''), more than the relations (none in this sample) --- which we hypothesized may be the reason our finetuned model performs better on \replace Attributes than Relations (c.f. Table \ref{tab:replace_split}). We separately generate more relation-targeted hard positives (with separate prompts to replace verbs and spatial prepositions), then sampling an equal number for relations and attributes, but the results when finetuning a model on this data did not differ significantly from those of our earlier finetuned model. Further study is required to improve model performance on \replace Relations.

\section{Finetuning on Hard Positives and Hard Negatives}
\label{sec:finetuning_details}

\begin{table*}[t]
\centering
\resizebox{0.7\textwidth}{!}{  
            \begin{tabular}{lccccccccc}
            \toprule
              \begin{tabular}{@{}l@{}}Mean $c$\\Score\end{tabular} & CLIP & \begin{tabular}{@{}l@{}}Neg-\\CLIP\end{tabular} & \begin{tabular}{@{}l@{}}CREPE\\-Swap\end{tabular} & \begin{tabular}{@{}l@{}}CREPE\\-Repl.\end{tabular} & SVLC & \begin{tabular}{@{}l@{}}SVLC\\+Pos\end{tabular} & \begin{tabular}{@{}l@{}}DAC\\-LLM\end{tabular} & \begin{tabular}{@{}l@{}}DAC\\-SAM\end{tabular} & Ours \\
              \midrule
              \texttt{REPL.} & 0.234 & 0.225 & 0.233 & 0.214 & 0.202 & 0.223 & 0.157 & 0.228 & 0.231\\
              \texttt{SWAP} & 0.255 & 0.239 & 0.250 & 0.228 & 0.211 & 0.228 & 0.132 & 0.224 & 0.247 \\
              \bottomrule \\
            \end{tabular}
        }
\caption{Mean image-text matching score of original caption $c$ per benchmark of all evaluated models. All hard negative-finetuned models reduce the image-text matching score of $c$, nearly all more so than our model finetuned on both hard negatives and hard positives.
}
\label{tab:mean_score}
\end{table*}
\subsection{Implementation details}
\label{sec:ft_implementation_details}
The finetuning follows the procedure outlined in SVLC \cite{doveh2023teaching}. For each training sample, one hard positive and one hard negative is retrieved and added to the batch.
The loss consists of: a contrastive loss across the batch, as in CLIP; a hard negative loss on each image with its original and negative captions; and a hard positive loss (called an analogy loss in SVLC) on each image with its original and positive captions. We finetune the model for 5 epochs on 4xA100 GPUs, which takes approximately 3 hours.

\subsection{Finetuning on both hard positives and hard negatives prevents reduction in model score of original caption}
\label{sec:prevent_reduction}
As discussed in Section \ref{sec:results} and \ref{sec:qual_analysis}, hard negative finetuning causes the model to award a lower image-text matching score to \textit{all} captions, not the hard negative caption alone. This has negative implications for various use cases where the score of the model is used directly, rather than as a ranking mechanism. 

Table \ref{tab:mean_score} shows the mean score awarded to the original caption $c$ by CLIP as well as various hard-negative finetuned models, showing that they all reduce the score of $c$ across both \replace and \swap (by $0.031$ on average). In comparison, our model, finetuned on both hard positives and hard negatives, reduces the score of the original caption much less (by $0.006$ on average) than all models except CREPE-Swap. CREPE-Swap assigns a higher score to $c$, but also an incorrectly higher score to $c_N$, resulting in much worse performance than our model on \swap and \replace (c.f. Table \ref{tab:results}). Our model strikes the best balance of high benchmark performance without significantly reducing the score of the original caption.

\section{Standard Evaluations}
\label{sec:standard_evals}
\setlength{\tabcolsep}{5pt}
\begin{table*}[t]
\centering
\resizebox{\textwidth}{!}{  

            \begin{tabular}{ll cc cc cc cc}
            \toprule
              
               &  & \multicolumn{2}{c}{ImageNet1k} &  \multicolumn{2}{c}{COCO} & \multicolumn{2}{c}{Flickr30k} & \multicolumn{2}{c}{VTAB}
               \\
               \cmidrule(lr){3-4} \cmidrule(lr){5-6} \cmidrule(lr){7-8} \cmidrule(lr){9-10}
               & Model & Acc@1 & Acc@5 & Image Recall@1 & Text Recall@1 & Image Recall@1 & Text Recall@1 & Acc@1 & Acc@5 \\
              \midrule
              \textcolor{gray}{(a)}&CLIP ViT-B/32& 63.33 & 88.83 & 30.46 & 50.14 & 58.82 & 77.40 & 39.00 & 70.90\\
              
              \midrule

              \textcolor{gray}{(b)}&CLIP-COCO & 53.18 & 81.98 & 50.34 & 66.76 & 68.48 & 83.40 & 34.67 & 68.55\\

              \midrule
              
              \textcolor{gray}{(c)} &Our HN & 50.40 & 79.58 & 49.61 & 63.98 & 67.80 & 80.10 & 32.40 & 67.53\\

               &Our HP+HN & 49.85 & 79.70 & 49.67 & 65.02 & 67.52 & 80.60 & 33.24 & 67.75 \\

              \bottomrule \\
            \end{tabular}
        }
\caption{Evaluation results on standard zero-shot tasks of (a) CLIP ViT-B/32, (b) CLIP ViT-B/32 finetuned on COCO train captions with neither hard positives nor hard negatives, (c) Our models. We report Acc@1 and Acc@5 for zero-shot classification on ImageNet1k and VTAB. For VTAB, we report the average over 20 zero-shot classification tasks~\cite{zhai2019large, ilharco_gabriel_2021_5143773}. For COCO and Flicker30k, we report Recall@1 for both image and text retrieval.
Comparing training with both hard positives and hard negatives (``Our HP + HN'') to training with hard negatives alone (``Our HN''), we see that we maintain --- or even improve --- performance on standard evaluation tasks, while improving model compositionality (c.f. Table~\ref{tab:results}).}
\label{tab:standard_eval}
\end{table*}


We conduct standard evaluations of our model on vision and vision-language tasks to ensure that our model did not experience catastrophic forgetting during finetuning. Table~\ref{tab:standard_eval} contains the results of our models evaluated on a wide range of zero-shot tasks. Specifically, we include zero-shot classification results on ImageNet-1K and 20 different VTAB tasks~\cite{zhai2019large}, as well as zero-shot retrieval performances on COCO and Flickr30k.
We include a CLIP model without finetuning, and a CLIP model finetuned on COCO alone (without hard positives or hard negatives) to serve as controlled baselines. \\

\noindent\textbf{Zero-shot classification performance drops.} From Table~\ref{tab:standard_eval}, we see that the models finetuned on the COCO training set show significant performance gains on COCO and Flickr30k retrieval, while losing performance on ImageNet-1K and VTAB classification tasks. This observation agrees with prior work~\cite{wortsman2022robust}, which shows that finetuning can decrease the robustness of CLIP models, particularly on different domains. Various methods have been proposed to effectively tackle the problem~\cite{wortsman2022robust, wortsman2022model}, and are orthogonal to this work.\\

\noindent\textbf{Adding hard positives improves compositionality while maintaining robustness, compared to training only with hard negatives.}
Comparing finetuning with hard positives and hard negatives to finetuning with hard negatives alone (as well as the COCO finetuning baseline with neither hard positives nor hard negatives), we see that adding hard positives to finetuning largely maintains the model's robustness on standard tasks while achieving significant improvements on compositionality.



\end{document}